\documentclass[10pt,twocolumn,letterpaper]{article}

\usepackage[pagenumbers]{cvpr} 
\usepackage{bm}
\usepackage{amsthm}
\newtheorem{fact}{Fact}  
\usepackage{multirow}
\usepackage{bold-extra}
\usepackage{pifont}
\usepackage{soul}  
\usepackage{graphicx}
\usepackage{wrapfig}
\usepackage{algorithm}
\usepackage{algpseudocode}
\usepackage{verbatim}

\definecolor{cvprblue}{rgb}{0.21,0.49,0.74}
\usepackage[pagebackref,breaklinks,colorlinks,allcolors=cvprblue]{hyperref}

\title{Mitigating Hallucinations in Large Vision-Language Models via DPO: \\
On-Policy Data Hold the Key}

\author{
Zhihe Yang$^{1,3}$\thanks{The work was conducted during the internship of Zhihe Yang (zhyang@link.cuhk.edu.hk) at Microsoft Research Asia.} \quad Xufang Luo$^{2}$\thanks{Corresponding author (xufang.luo@microsoft.com)} \quad Dongqi Han$^{2}$ \quad Yunjian Xu$^{1,3}$\thanks{Corresponding author (yjxu@mae.cuhk.edu.hk)} \qquad Dongsheng Li$^{2}$ \\
$^{1}$The Chinese University of Hong Kong, Hong Kong SAR, China\\
$^{2}$Microsoft Research Asia, Shanghai, China\\
$^{3}$The Chinese University of Hong Kong, Shenzhen Research Institute (SZRI), Guangdong, China\\
\textcolor{VioletRed}{\url{https://opa-dpo.github.io}}\vspace{-2mm}
}

\begin{document}
\maketitle
\vspace{-2mm}
\begin{abstract}
Hallucination remains a major challenge for Large Vision-Language Models (LVLMs).
Direct Preference Optimization (DPO) has gained increasing attention as a simple solution to hallucination issues. It directly learns from constructed preference pairs that reflect the severity of hallucinations in responses to the same prompt and image.
Nonetheless, different data construction methods in existing works bring notable performance variations. 
We identify a crucial factor here: outcomes are largely contingent on whether the constructed data aligns on-policy w.r.t the initial (reference) policy of DPO.
Theoretical analysis suggests that learning from off-policy data is impeded by the presence of KL-divergence between the updated policy and the reference policy. 
From the perspective of dataset distribution, we systematically summarize the inherent flaws in existing algorithms that employ DPO to address hallucination issues. 
To alleviate the problems, we propose On-Policy Alignment (OPA)-DPO framework, which uniquely leverages expert feedback to correct hallucinated responses and aligns both the original and expert-revised responses in an on-policy manner. 
Notably, with only 4.8k data, OPA-DPO achieves an additional reduction in the hallucination rate of  LLaVA-1.5-7B: 13.26\% on the AMBER benchmark and 5.39\% on the Object-Hal benchmark, compared to the previous SOTA algorithm trained with 16k samples. Our implementation is available at \textcolor{VioletRed}{\url{https://github.com/zhyang2226/OPA-DPO}}.
\end{abstract}    
\vspace{-4mm}
\section{Introduction}

\label{sec:intro}

\begin{figure}[t]
  \centering
  \vspace{-3mm}
   \includegraphics[width=1.0\linewidth]{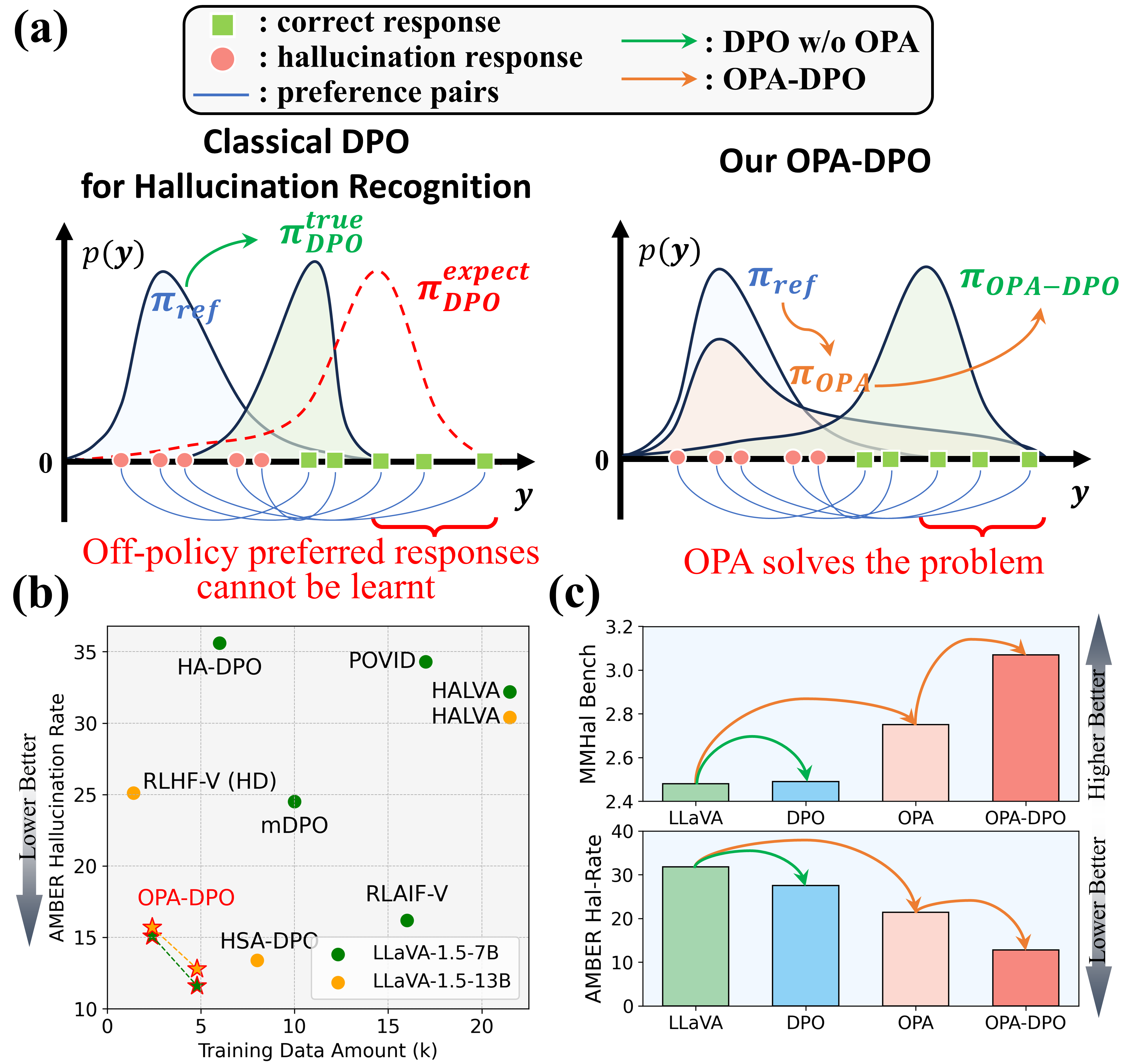}
   \vspace{-5mm}
   \caption{(a) \textbf{OPA-DPO motivation:} Naive adoption of DPO struggles to learn off-policy preferred responses due to the substantial reverse KL-divergence constraint (induced by unmatched supports). Our OPA operation aligns these responses on-policy, enabling effective learning with subsequent DPO. (b) \textbf{Data scale vs. performance:} We present the AMBER hallucination rates for various DPO-based algorithms and their training data volume. OPA-DPO (star markers) achieves SOTA performance with minimal amount of data. (c) \textbf{Impact of OPA:} Using LLaVA-1.5-13B with 4.8k data, we evaluate performance of DPO with/without OPA operations. The inclusion of OPA significantly enhances performance compared to DPO alone.}\label{fig1_overall}
   \vspace{-5mm}
\end{figure}

Recent advancements in instruction-following Large Vision-Language Models (LVLMs) have achieved significant milestones \cite{alayrac2022flamingo, dai2023instructblip, li2023blip2, openai2023gpt4v}. By integrating pre-trained vision encoders with Large Language Models (LLMs) and fine-tuning them on instruction-based datasets, the combined models demonstrate remarkable image understanding capabilities \cite{liu2024llava, liu2024llava2, bai2023qwenvl}. This technology shows considerable potential across various fields, including image captioning \cite{liu2024llava2}, pathology recognition \cite{li2024llavamed, sun2024drllava}, and medical imaging diagnostic \cite{hyland2023maira, bannur2024maira2}. 
Nevertheless, a significant barrier hinders their practical application: hallucinations \cite{bai2024hal_survey, lan2024survey, liu2024survey}, 
which refer to discrepancies between the image's actual content and the model-generated text.
Such issue is pronounced in LVLMs.
Even the most advanced GPT-4V \cite{openai2023gpt4v} exhibits hallucinations in 45.9\% of responses for certain tasks \cite{yu2024rlhfv}.

\begin{figure*}[t]
  \centering
   \includegraphics[width=0.9\linewidth]{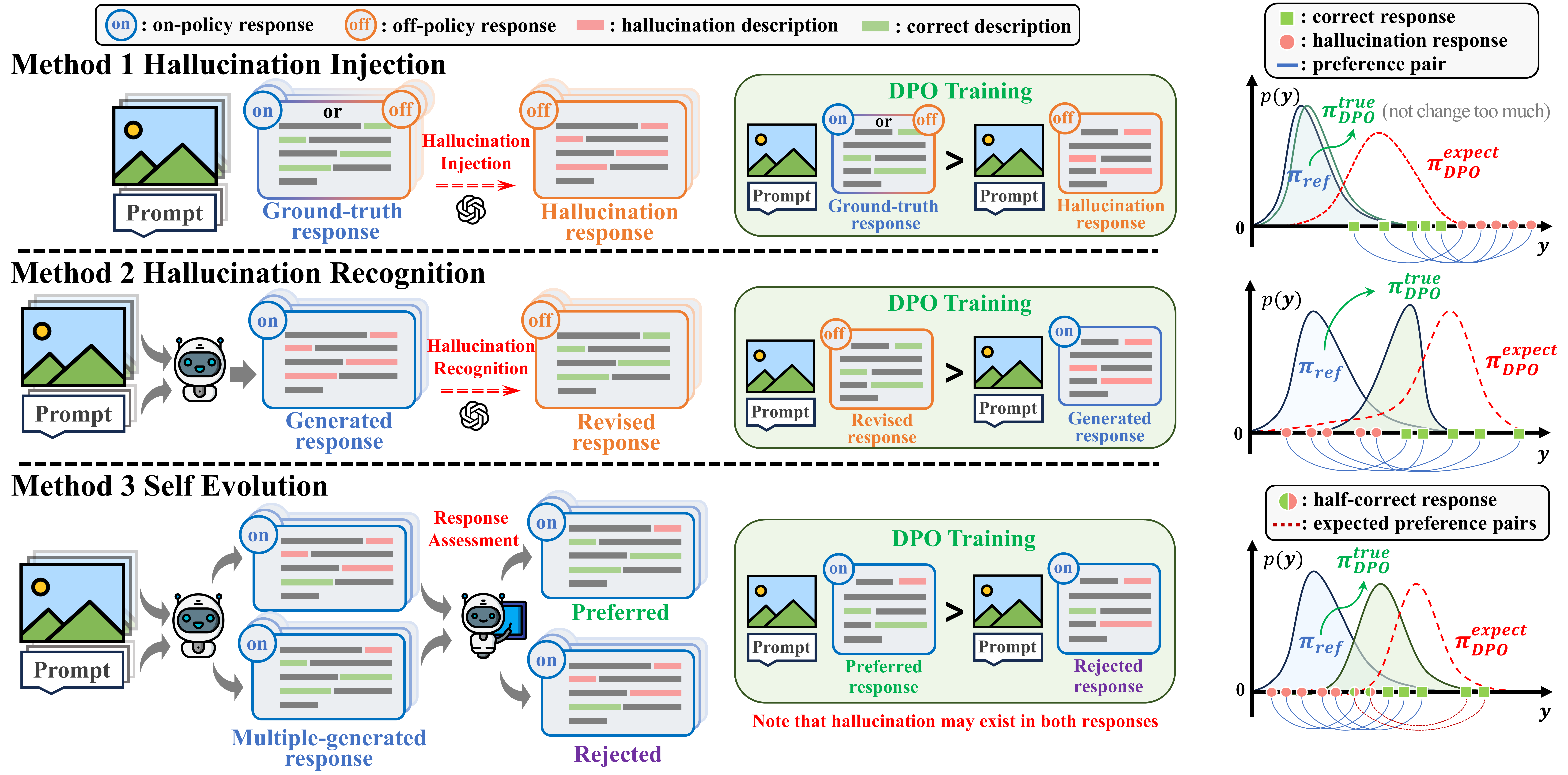}
   \vspace{-3mm}
   \caption{We categorize existing DPO-based algorithms for addressing hallucination issues in LVLMs into 3 classes: (1) \textbf{Hallucination Injection} (POVID \cite{zhou2024povid} and HALVA \cite{sarkar2024halva}). The ground-truth response is preferred, while the rejected response contains injected hallucinations. Since the errors do not originate from the model itself, the policy is unlikely to benefit from training. (2) \textbf{Hallucination Recognition} (RLHF-V \cite{yu2024rlhfv}, HA-DPO \cite{zhao2023hadpo} and HSA-DPO \cite{xiao2024hsadpo}). 
   The model generates responses, after which experts (AI or human) identify errors and make revisions. The off-policy nature of the revised responses makes them challenging to learn effectively.
   (3) \textbf{Self Evolution} (RLAIF-V \cite{yu2024rlaif}). Both preferred and rejected responses are generated by the initial policy. A superior model assesses hallucinations, preferring the response with fewer errors. However, hallucinations may exist in both responses, thereby affecting the learning efficiency.}\label{fig2_DPO_sum}
   \vspace{-4mm}
\end{figure*}

Among numerous studies aimed at reducing hallucinations in LVLMs, a remarkable approach is further fine-tuning the models using Reinforcement Learning from Human Feedback (RLHF) \cite{yu2024rlhfv, sun2023llava_rlhf} or AI Feedback (RLAIF) \cite{xiao2024hsadpo, zhao2023hadpo, yu2024rlaif}. 
RLH(AI)F aligns the model to the correct direction by fine-tuning it with constructed preference pairs, where the win response has less hallucination to the identical image and prompt than the loss one.
RLH(AI)F algorithms can be broadly categorized into two classes: Proximal Policy Optimization (PPO) \cite{schulman2017ppo} and Direct Preference Optimization (DPO) \cite{rafailov2024dpo}. DPO is generally simpler than PPO in practice because it streamlines the operational framework by eliminating the need for reward model training and the online rollout process in dataset generation, but relying solely on pre-collected offline data.

Indeed, PPO \cite{schulman2017ppo} and DPO \cite{rafailov2024dpo} share the same learning objective: to maximize the reward derived from the Bradley-Terry model \cite{bradley1952btmodel} while constraining the Kullback-Leibler (KL) divergence \cite{kullback1951kldiv} between the updated policy and the initial (reference) policy.
However, a key distinction arises during the training process: PPO, an online and on-policy algorithm, requires the use of online rollout data, whereas DPO relies entirely on offline datasets, which may be collected by any policy in practice \cite{li2023silkie, wang2024mdpo}.
Therefore, owing to the constraint of reverse KL-divergence, the data used in PPO training process is predominantly 
on-policy\footnote{In this paper, \textit{on-policy} denotes data with high sampling probability under the initial policy, while \textit{off-policy} indicates low or near-zero.}
in relation to the initial (reference) policy.
In contrast, for DPO, where policy is trained completely offline, the alignment of training data with the reference policy is rarely considered, especially for LVLM works.

In this paper, we reveal a key insight: \textit{the on-policy property of training data, which was neglected by DPO-based algorithms used to train LVLMs, plays a crucial role in enabling effective DPO training}.
As illustrated in Figure~\ref{fig1_overall}a left, strictly off-policy preferred responses cannot be learned by naive DPO. The limitation stems from the fact that assigning even a small positive probability to such off-policy data leads to substantially large KL-divergence between updated policy and reference policy, due to the mismatches in their support (see detailed analysis in Chapter~\ref{sec:analysis}).

Based on the on-policy property, we classify existing algorithms that employ DPO to tackle the hallucination issues for LVLMs into three distinct categories, as shown in Figure~\ref{fig2_DPO_sum}. 
Despite minor differences in datasets and training parameters, Method 3 significantly outperforms the other two methods. This can be attributed to Method 3's exclusive use of on-policy preference pairs, whereas the other two methods involve either preferred or rejected responses off-policy.
Notably, Method 3 has an inherent limitation: persistent hallucinations may exist in both preferred and rejected responses, since all responses are generated by the policy to be updated.
This shortcoming results in inefficient learning and the requirement on a large volume of data to achieve satisfying performance.

Converting these insights into solutions, we propose a novel framework: On-Policy Alignment (OPA)-DPO (cf. Figure~\ref{fig1_overall}a right), which ensures that data remains on-policy while simultaneously leveraging expert guidance to improve learning efficiency of LVLMs. 
We first utilize GPT-4V \cite{openai2023gpt4v} to recognize hallucination and deliver fine-grained revisions to the model-generated responses.
Then we align these off-policy adjustments \textit{on-policy} through fine-tuning the initial policy. 
The operation allows the subsequent DPO training to circumvent the constraints imposed by KL-divergence and effectively incorporate these changes.

Our contributions are threefold: (1) We identify an intrinsic property of DPO: its high reliance on on-policy data. (2) We summarize the inherent flaws of existing algorithms that employ DPO to address the hallucination problem. (3) Building upon the identified shortcomings of existing methods, we propose OPA-DPO, a novel framework that utilizes 4.8k data to achieve state-of-the-art (SOTA) performance on hallucination benchmarks, surpassing previous methods relying on larger datasets (Figure~\ref{fig1_overall}b,c).

\section{Preliminary}
\label{sec:pre}
\paragraph{Large Vision-Language Models.}
LVLMs represent a class of multimodal models that integrate visual and linguistic information to generate outputs in natural language \cite{liu2024survey}. 
Typically, LVLMs comprise three components \cite{liu2024llava}: a visual encoder, a modality connection module, and an LLM. The visual encoder transforms input images ($\mathbf{m}$) into visual tokens. The connection module aligns these visual tokens with the LLM's word embedding space. Combined with a user-provided linguistic prompt ($\mathbf{x}$), the LLM generates response ($\mathbf{y}$) in an auto-regressive manner.
\vspace{-3mm}
\paragraph{Direct Preference Optimization.}
To further enhance the performance of LVLMs, RLHF/RLAIF necessitates a reward model $r(\mathbf{x}, \mathbf{y}, \mathbf{m})$, which evaluates human preferences for the response $\mathbf{y}$ given the prompt $\mathbf{x}$ and the image $\mathbf{m}$. The fundamental learning objective is expressed as
\vspace{1mm}
{\small
\begin{equation}\label{eq1}
    \max\nolimits_{\pi_\theta} \mathbb{E}_{\mathcal{D}}[r(\mathbf{x},\mathbf{y},\mathbf{m})] - \\ \beta\mathbb{D}_{\mathrm{KL}}[\pi_\theta( \cdot |\mathbf{x},\mathbf{m}) || \pi_{\mathrm{ref}}( \cdot |\mathbf{x},\mathbf{m})],
\end{equation}}
\vspace{1mm}
where $\mathcal{D}$ represents the datasets where the prompts and images are sampled from. $\mathbb{D}_{\mathrm{KL}}$ stands for the KL-divergence, 
and $\beta$ controls the degree of regularization.
DPO \cite{rafailov2024dpo} derives the closed-form optimal solution for Eq.~\eqref{eq1} and identify the reward function can be analytically expressed via
\vspace{0mm}
{\small
\begin{equation}\label{eq2}
    r(\mathbf{x}, \mathbf{y}, \mathbf{m})=\beta\log\frac{\pi_\theta(\mathbf{y}| \mathbf{x}, \mathbf{m})}{\pi_\mathrm{ref}(\mathbf{y}|\mathbf{x},\mathbf{m})} + \beta \log Z(\mathbf{x},\mathbf{m}),
\end{equation}}
\vspace{1mm}
where $Z(\mathbf{x}, \mathbf{m})$ is a partition function that only depends on prompt $\mathbf{x}$ and image $\mathbf{m}$. Incorporating with Bradley-Terry model \cite{bradley1952btmodel}, and the dataset comprising preference pairs ($\mathbf{y}_w$ over $\mathbf{y}_l$) towards the same prompt $\mathbf{x}$ and image $\mathbf{m}$, the model can be directly optimized through
\vspace{1mm}
{\small
\begin{equation}\label{eq3}
\begin{aligned}
    & \mathcal{L}_\mathrm{DPO} = -\mathbb{E}_\mathcal{D} [\log \sigma(r(\mathbf{x},\mathbf{y}_w,\mathbf{m}) -r(\mathbf{x},\mathbf{y}_l,\mathbf{m}))] \\
    & =-\mathbb{E}_\mathcal{D} [\log \sigma(
    \scalebox{1.05}{$\beta\log \frac{\pi_\theta(\mathbf{y}_w|\mathbf{x},\mathbf{m})}{\pi_\mathrm{ref}(\mathbf{y}_w|\mathbf{x},\mathbf{m})} \! - \!\beta\log\frac{\pi_\theta(\mathbf{y}_l|\mathbf{x},\mathbf{m})}{\pi_\mathrm{ref}(\mathbf{y}_l|\mathbf{x},\mathbf{m})} $} )],
\end{aligned}
\end{equation}}
\vspace{1mm}
where $\sigma(\cdot)$ denotes the sigmoid function.
\vspace{-3mm}
\paragraph{Supervised Fine Tuning.}
As the most commonly used technique for LLMs and multimodal LLMs, Supervised Fine Tuning (SFT) is a simple and efficient method to align pre-trained models with downstream tasks. 
Given a dataset $\mathcal{D}$ including prompts $\mathbf{x}$, images $\mathbf{m}$, and the corresponding standard responses $\mathbf{y}$, the training loss for SFT is
{\small
\begin{equation}\label{eq4}
    \mathcal{L}_\mathrm{SFT} = - \mathbb{E}_\mathcal{D}\left[ \sum^L_i \sum^C_c \mathbb{I}(y_i^c) \log \pi_\theta(y_i^c | \mathbf{x}, \mathbf{m}, \mathbf{y}_{<i}) \right],
\end{equation}}
where $L$ is the length of the response, $C$ is the number of possible classes or tokens, $\mathbb{I}(y_i^c)$ is an indicator function that equals 1 if the $i$-th token is of class $c$ and 0 otherwise, and $\pi_\theta(y_i^c | \mathbf{x}, \mathbf{m}, \mathbf{y}_{<i})$ represents the model's predicted probability of the $i$-th token given the prompt $\mathbf{x}$, image $\mathbf{m}$, and the sequence of preceding tokens $\mathbf{y}_{<i}$. 
\vspace{-3mm}
\paragraph{On-Policy Data.}
In the realm of reinforcement learning (RL), on-policy data is sampled from the current policy and becomes off-policy after the policy is updated \cite{levine2020offline}. As a fine-tuning process, policy updates for LLMs do not significantly change its sampling probabilities. 
In this paper, we define a response $\mathbf{y}$ to a prompt $\mathbf{x}$ and image $\mathbf{m}$ as on-policy if $\pi_\mathrm{ref}(\mathbf{y}|\mathbf{x}, \mathbf{m}) > \epsilon$, where $\epsilon$ is a small positive threshold.
\section{Problem Analysis}
\label{sec:analysis}
Three questions reflect our thinking path in this work: 
\begin{itemize}[label=$\bullet$]
    \item \textbf{Q1}: How does the dataset distribution relative to the initial/reference policy affect the performance of DPO? 
    \item \textbf{Q2}: What are the inherent flaws of other algorithms that employ DPO to tackle hallucination problems? 
    \item \textbf{Q3}: What adjustments can be made to current frameworks to rectify their intrinsic deficiencies?
\end{itemize}  
\vspace{-3mm}
\paragraph{Question 1}
Note that the minimizer of Eq.~\eqref{eq3} corresponds to the optimal solution for Eq.~\eqref{eq1}. Nevertheless, by reconsidering the definition of the KL-divergence
{\small
\begin{equation}\label{eq5}
    \mathbb{D}_{\mathrm{KL}}[P \Vert Q] := \sum\nolimits_{y\in\mathcal{Y}}P(y) log \scalebox{1.05}{$\frac{P(y)}{Q(y)}$},
\end{equation}}
where $P$ and $Q$ represent two distinct probability distributions. We can deduce the following fact
\vspace{-1.5mm}
\begin{fact}\label{fact1}
    Given a prompt $\mathbf{x}$ and an image $\mathbf{m}$, suppose there exists one response $\mathbf{y}$ such that $\pi_\theta(\mathbf{y}|\mathbf{x},\mathbf{m})>0$, whereas $\pi_\mathrm{ref}(\mathbf{y}|\mathbf{x},\mathbf{m}) \rightarrow 0$, the KL-divergence between the two policy has $\mathbb{D}_{\mathrm{KL}}[\pi_\theta( \cdot |\mathbf{x},\mathbf{m}) || \pi_{\mathrm{ref}}( \cdot |\mathbf{x},\mathbf{m})] \rightarrow \infty$.
\end{fact} 
\vspace{-1.5mm}

Fact~\ref{fact1} illustrates that if the preferred response has a near-zero probability with respect to the initial/reference policy, i.e., it is strictly off-policy data, then it can \textbf{never} be learned by any policy that begins from the learning objective outlined in Eq.~\eqref{eq1}.
In other words, denoting the support for the updated policy as $\mathcal{Y}_\theta$, the support for the initial policy as $\mathcal{Y}_\mathrm{ref}$, and the global sampling space as $\mathcal{Y}_\mathrm{global}$, we always have the relationship $\mathcal{Y}_\theta \subseteq \mathcal{Y}_\mathrm{ref} \subseteq \mathcal{Y}_\mathrm{global}$. Any responses falling into the set $\mathcal{Y}_\mathrm{global} \setminus \mathcal{Y}_\mathrm{ref}$ cannot be learned by $\pi_\theta$. It should be noted that this issue arises only with DPO, as PPO samples responses on-line from $\mathcal{Y}_\theta$.

A natural question arises: how does DPO deal with these off-policy preferred responses? By taking the partial derivative of Eq.~\eqref{eq3}, the gradient with respect to the policy parameters $\theta$ can be expressed as

\vspace{-2mm}
{\footnotesize
\begin{equation}\label{eq6}
\begin{aligned}
    \nabla_\theta \mathcal{L}_\mathrm{DPO} 
    & = - \mathbb{E}_{(\mathbf{y}_w, \mathbf{y}_l, \mathbf{x}, \mathbf{m} ) \sim \mathcal{D}} \\
     \bigg[ \beta \cdot \sigma &\bigg( - \underbrace{\beta \log \frac{\pi_\theta(\mathbf{y}_w|\mathbf{x},\mathbf{m})}{\pi_\mathrm{ref}(\mathbf{y}_w|\mathbf{x},\mathbf{m})}}_{r_w} +
    \underbrace{\beta \log \frac{\pi_\theta(\mathbf{y}_l|\mathbf{x},\mathbf{m})}{\pi_\mathrm{ref}(\mathbf{y}_l|\mathbf{x},\mathbf{m})}}_{r_l} \bigg) \cdot  \\
    & \big(
    \underbrace{\nabla_\theta \log \pi_\theta(\mathbf{y}_w|\mathbf{x},\mathbf{m})-\nabla_\theta \log \pi_\theta(\mathbf{y}_l|\mathbf{x},\mathbf{m})}_{\mathrm{log-likelihood}}
    \big)
    \bigg],
\end{aligned}
\end{equation}}
where $\sigma(r_w-r_l)=0.5$ prior to the policy update. Assuming that the preferred response $\mathbf{y}_w$ is off-policy, $\pi_\theta(\mathbf{y}_w|\mathbf{x},\mathbf{m})$ undergoes a single step of log-likelihood maximization with a coefficient of $0.5\beta$. Following this update, $r_w$ inclined to become substantially large, thereby causing $\sigma(r_w-r_l) \rightarrow 0$. Nonetheless, the increment in probability induced by this single-step update proves insufficient for the preferred response to be sampled during the auto-regressive generation process.
In summary, the low likelihood of off-policy preferred responses (relative to reference policy) drives the DPO updating weight toward zero, thereby rendering effective learning nearly impossible.
\vspace{-4mm}
\paragraph{Question 2}
For algorithms adopting Method 1 (Hallucination Injection) outlined in Figure~\ref{fig2_DPO_sum}, a ground-truth (GT) response is deemed on-policy if it has been incorporated into fine-tuning SFT dataset. The shortcoming is evident: hallucinations do not originate from the model itself. While the probability associated with the GT response is augmented, the probability of model-intrinsic hallucinations is neither explicitly identified nor substantially diminished.

Method 2, Hallucination Recognition, is the most widely adopted approach, with the majority of studies opting to use GPT-4 with ground-truth image captions or GPT-4V as the recognizer. 
Nevertheless, a significant challenge persists: the preferred response often remains off-policy, as highlighted in our answers to Question 1. 

Method 3, Self Evolution, is exclusively employed by RLAIF-V \cite{yu2024rlaif}, which significantly outperforms the previous two methods in hallucination benchmarks. However, it has a notable shortcoming: since the method relies on the model generating two responses to form a preference pair, it cannot effectively address intrinsic hallucinations present in both responses. As a result, this approach requires a substantial amount of data and multiple iterative updates.
\vspace{-4mm}
\paragraph{Question 3}
Method 2 employs domain experts to construct preferred responses, establishing a robust paradigm but encountering the off-policy issue. Although Method 3 addresses this challenge, the reliability of the preferred responses is compromised. 
To synergize the strengths of both approaches, namely aligning expert-revised preferred responses with the on-policy framework, it is essential to consider modifications to the model itself before commencing DPO training.
A promising method that comes to mind is adapting Low Rank Adaptation (LoRA) SFT to the expert revision, which our experimental evidence demonstrates to be exceptionally effective.
In conjunction with our adjusted DPO training loss, OPA-DPO is capable of achieving SOTA performance with a minimal data requirement. 
\section{On-Policy Alignment DPO}\label{sec_OPADPO}

\begin{figure}[t]
  \centering
   \includegraphics[width=1.0\linewidth]{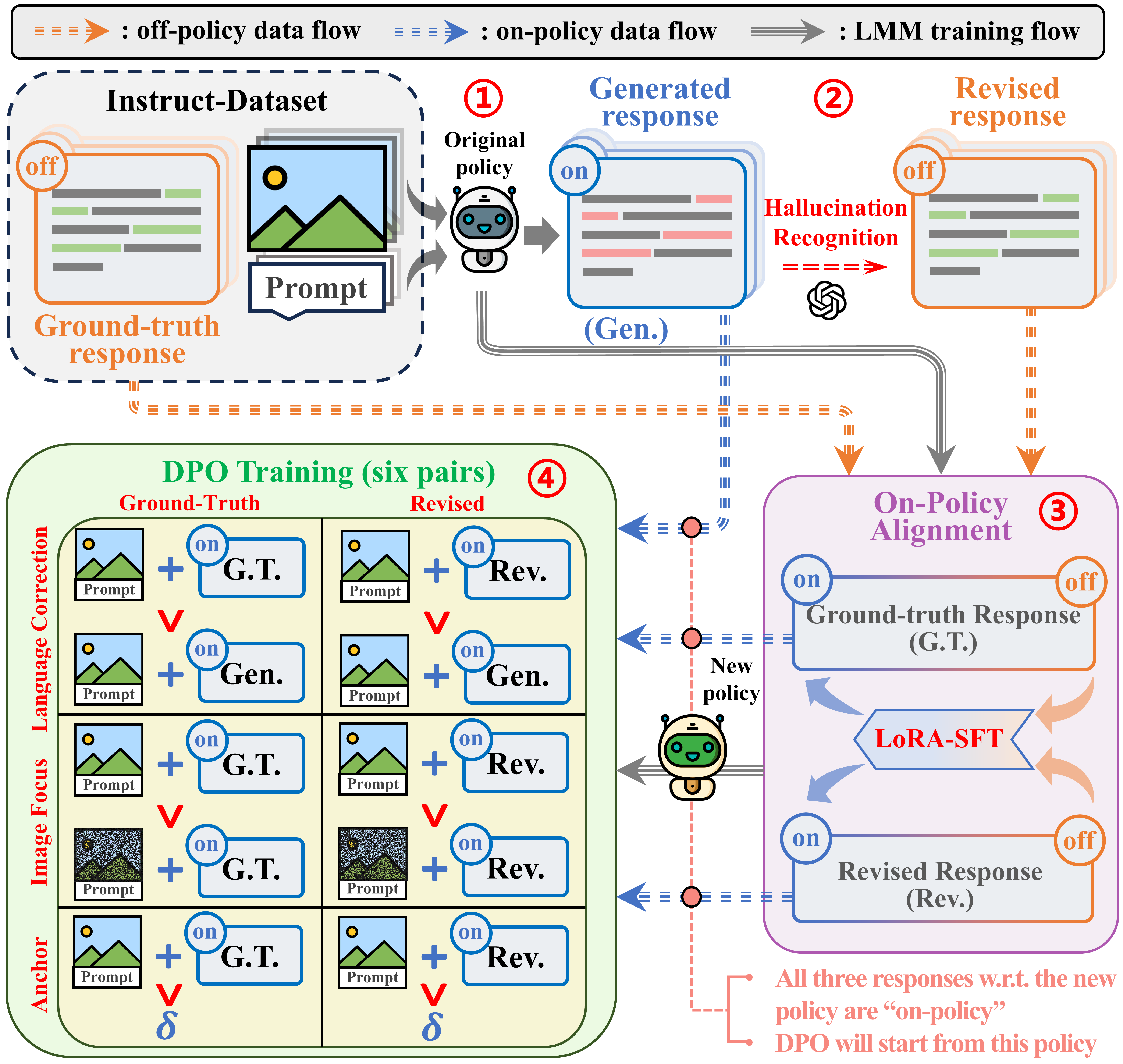}
   \caption{Our proposed OPA-DPO comprises four essential steps: \ding{172} Collect responses from the original policy based on the images and corresponding prompts. \ding{173} Utilize GPT-4V to correct any hallucinations in the generated responses with minimal modifications. \ding{174} Conduct LoRA-SFT on the GT responses and revised responses. \ding{175} Initiate OPA-DPO training from the policy obtained in step 3.}
   \label{fig3_OPA-DPO}
   \vspace{-3mm}
\end{figure}

As illustrated in Figure~\ref{fig3_OPA-DPO}, our proposed OPA-DPO framework encompasses four essential steps. The initial two steps, designated as data collection, along with the third step, on-policy alignment, are detailed in Chapter~\ref{pre_OPA_DPO}. The final step, OPA-DPO training, is elaborated in Chapter~\ref{OPA_DPO}.

\subsection{Data Collection and On-Policy Alignment}\label{pre_OPA_DPO}
Initially, we instruct the model (slated for training) to generate responses based on pre-collected images and prompts, using a combination of ``top-k" and ``top-p" sampling methods. 
Following this, 
we supply GPT-4V with the generated responses $\mathbf{y}_\mathrm{Gen}$, the original prompts $\mathbf{x}$, the images $\mathbf{m}$, and the GT responses $\mathbf{y}_\mathrm{GT}$. 
GPT-4V is tasked with identifying hallucinations by evaluating the generated responses at the sentence level. 
Each sentence within a response is assigned a score, $S_\mathrm{hal}$, which indicates the severity of the hallucination. 
Moreover, GPT-4V is required to categorize sentences with incorrect description as either \textit{image recognition errors} or \textit{language comprehension errors}, with the classification results represented by $S_\mathrm{img}$. 
Additionally, GPT-4V is also instructed to make minimal revisions to any erroneous sentences, and the aggregate of these revised sentences is denoted as $\mathbf{y}_\mathrm{Rev}$ (refer to Appendix for further details).

Subsequently, we integrate the GT responses with the GPT-4V revised responses to construct an instruction-following dataset, which includes $\mathbf{y}_\mathrm{GT}, \mathbf{y}_\mathrm{Rev}, \mathbf{x}, \mathbf{m}$. 
We then perform LoRA-SFT on this dataset, utilizing the loss function in Eq.~\eqref{eq4}. We denote the resulting policy from this phase as $\pi_\mathrm{OPA}$. Note that this policy serves as the reference (initial) policy for the subsequent OPA-DPO training.

\subsection{OPA-DPO Training}\label{OPA_DPO}
Compared to classical DPO, which forms only a single language preference pair, our OPA-DPO loss comprises three distinct components, each containing two pairs:

\vspace{-4mm}\normalsize
\paragraph{Language Corrections.}
As the most basic component of DPO, the language-level preference is naturally formed between the GT response and the generated response, as well as between the revised response and the generated response.
Following the approach outlined in RLHF-V \cite{yu2024rlhfv}, we aim to concentrate the policy update on the erroneous sections and their respective corrections.
To achieve this, we construct a mapping from the GPT-4V marked hallucination scores to establish the update weight $W_\mathrm{hal}(S_\mathrm{hal})$. Then the hallucination-weighted log-policy is defined as $\log \pi^\mathrm{hw}(\mathbf{y}|\mathbf{x}, \mathbf{m})= \sum_{i}^{L} W_\mathrm{hal}(S^i_\mathrm{hal}) \log\pi(y_i|\mathbf{x}, \mathbf{m}, \mathbf{y}_{<i})$, where $L$ represents the response length, and $S^i_\mathrm{hal}$ denotes the hallucination score for token $y_i$. Note that $S^i_\mathrm{hal}$ remain consistent for tokens within the same sentence but may vary between different sentences. Subsequently, we can form language correction preference pairs

\vspace{-2mm}
{\footnotesize
\begin{equation}\label{eq7}
\begin{aligned}
    & \mathcal{L}_\mathrm{LC} = - \mathbb{E}_{(\mathbf{y}_\mathrm{GT}, \mathbf{y}_\mathrm{Rev}, \mathbf{y}_\mathrm{Gen}, \mathbf{x}, \mathbf{m}) \sim \mathcal{D}} \\
    & \bigg[\log \sigma \bigg(
    \beta\log \tfrac{\pi_\theta(\mathbf{y}_\mathrm{GT}|\mathbf{x},\mathbf{m})}{\pi_\mathrm{OPA}(\mathbf{y}_\mathrm{GT}|\mathbf{x},\mathbf{m})} -\beta\log \tfrac{\pi_\theta(\mathbf{y}_\mathrm{Gen}|\mathbf{x},\mathbf{m})}{\pi_\mathrm{OPA}(\mathbf{y}_\mathrm{Gen}|\mathbf{x},\mathbf{m})} \bigg) \\
    & + \log \sigma \bigg(
    \beta\log 
    \tfrac{\pi_\theta^\mathrm{hw}
    (\mathbf{y}_\mathrm{Rev}|\mathbf{x},\mathbf{m})}
    {\pi_\mathrm{OPA}^\mathrm{hw}(\mathbf{y}_\mathrm{Rev}|\mathbf{x},\mathbf{m})} 
    - \beta\log \tfrac{\pi_\theta^\mathrm{hw}(\mathbf{y}_\mathrm{Gen}|\mathbf{x},\mathbf{m})}{\pi_\mathrm{OPA}^\mathrm{hw}(\mathbf{y}_\mathrm{Gen}|\mathbf{x},\mathbf{m})} 
    \bigg)
    \bigg].
\end{aligned}
\end{equation}}
\vspace{-3.2mm}
\paragraph{Image Focus Mechanism.}
A critical obstacle in ensuring LVLMs properly engage with images is their innate tendency to ignore the visual modality during the optimization phase \cite{wang2024mdpo, zhou2024povid}. 
Intuitively, when the image data is compromised, the probability that the model produces the correct response diminishes.
Building upon mDPO \cite{wang2024mdpo}, we form preference pairs between the original images $\mathbf{m}$ and distorted images $\mathbf{m'}$, using the same prompts and GT/revised responses.
Furthermore, we expect this mechanism to be more effective for sentences where understanding of the image itself is biased.
To accomplish this, we create another mapping from the GPT-4V marked categorization results $S_\mathrm{img}$ to determine the update weight $W_\mathrm{img}(S_\mathrm{img})$. We then describe the image-weighted log-policy as 
$\log \pi^\mathrm{iw}(\mathbf{y}|\mathbf{x}, \mathbf{m})= \sum_{i}^{L} W_\mathrm{img}(S^i_\mathrm{img}) \log\pi(y_i|\mathbf{x}, \mathbf{m}, \mathbf{y}_{<i})$. This allows us to subsequently establish image focus preference pairs

\vspace{-2mm}
{\footnotesize
\begin{equation}\label{eq8}
\begin{aligned}
    & \mathcal{L}_\mathrm{IF} = - \mathbb{E}_{(\mathbf{y}_\mathrm{GT}, \mathbf{y}_\mathrm{Rev}, \mathbf{x}, \mathbf{m}, \mathbf{m'}) \sim \mathcal{D}} \\
    & \bigg[ \log \sigma \bigg(
    \beta\log \tfrac{\pi_\theta(\mathbf{y}_\mathrm{GT}|\mathbf{x},\mathbf{m})}{\pi_\mathrm{OPA}(\mathbf{y}_\mathrm{GT}|\mathbf{x},\mathbf{m})} -\beta\log \tfrac{\pi_\theta(\mathbf{y}_\mathrm{GT}|\mathbf{x},\mathbf{m'})}{\pi_\mathrm{OPA}(\mathbf{y}_\mathrm{GT}|\mathbf{x},\mathbf{m'})} \bigg) \\
    & + \log \sigma \bigg(
    \beta\log 
    \tfrac{\pi_\theta^\mathrm{iw}
    (\mathbf{y}_\mathrm{Rev}|\mathbf{x},\mathbf{m})}
    {\pi_\mathrm{OPA}^\mathrm{iw}(\mathbf{y}_\mathrm{Rev}|\mathbf{x},\mathbf{m})} 
    - \beta\log \tfrac{\pi_\theta^\mathrm{iw}(\mathbf{y}_\mathrm{Rev}|\mathbf{x},\mathbf{m'})}{\pi_\mathrm{OPA}^\mathrm{iw}(\mathbf{y}_\mathrm{Rev}|\mathbf{x},\mathbf{m'})} 
    \bigg)
    \bigg].
\end{aligned}
\end{equation}}

\vspace{-4mm}
\paragraph{Anchored Preference.}
Numerous studies \cite{xu2024dpo_ppo, wang2024mdpo, zhao2023hadpo} document a reduced likelihood of preferred response during the DPO training process. This trend may be attributed to the intrinsic characteristics of DPO, which concentrates on the relative preferences. Our findings align with these studies and we observe that the reduction
adversely affects downstream performance. Following mDPO \cite{wang2024mdpo}, we employ two anchors to constrain the preferred

\vspace{-2mm}
{\footnotesize
\begin{equation}\label{eq9}
\begin{aligned}
    \mathcal{L}_\mathrm{Anc} \!= \!- \mathbb{E}_{(\mathbf{y}_\mathrm{GT}, \mathbf{y}_\mathrm{Rec}, \mathbf{x}, \mathbf{m}) \sim \mathcal{D}} &
    \bigg[ \log \sigma \bigg(
    \beta\log \tfrac{\pi_\theta(\mathbf{y}_\mathrm{GT}|\mathbf{x},\mathbf{m})}{\pi_\mathrm{OPA}(\mathbf{y}_\mathrm{GT}|\mathbf{x},\mathbf{m})} \! - \! \delta \bigg) \\
    +  \log \sigma & \bigg(
    \beta\log 
    \tfrac{\pi_\theta
    (\mathbf{y}_\mathrm{Rec}|\mathbf{x},\mathbf{m})}
    {\pi_\mathrm{OPA}(\mathbf{y}_\mathrm{Rec}|\mathbf{x},\mathbf{m})} 
    -  \delta
    \bigg)
    \bigg].
\end{aligned}
\end{equation}}
Combining Eqs.~\eqref{eq7}\eqref{eq8}\eqref{eq9}, we get the loss for OPA-DPO:

\vspace{-2mm}
{\footnotesize
\begin{equation}\label{eq10}
    \mathcal{L}_\mathrm{OPA-DPO} = \mathcal{L}_\mathrm{LC} + \gamma_1 \mathcal{L}_\mathrm{IF} + \gamma_2 \mathcal{L}_\mathrm{Anc}.
\end{equation}}
It should be noted that each component is crucial and cannot be omitted. See Chapter~\ref{sec_ablation} for ablation studies.
\section{Experiments}\label{sec5_exp}

\subsection{Experimental Setup}\label{sec_exp_details}

\paragraph{Models and Datasets.}
We apply OPA-DPO on two LVLMs with distinct parameter sizes: LLAVA-v1.5-7B and LLAVA-v1.5-13B, both using CLIP ViT-L-336px as the vision encoder. The 7B model is based on Vicuna-7B, and the 13B on Vicuna-13B. Each model underwent pretraining on 558K image-text pairs and was subsequently fine-tuned on 665K instruction-based samples. As for the datasets, we randomly selected 4.8K samples from the RLAIF-V \cite{yu2024rlaif} datasets, using their preferred response as the ground truth.

\vspace{-3mm}
\paragraph{Evaluation Metrics.}
We focuses on mitigating hallucinations in LVLMs, with experiments conducted on four benchmarks:
\textbf{1) AMBER} \cite{wang2023amber} A benchmark with detailed object annotations, featuring 1004 images in a generative task. Using the official codebase, we evaluate CHAIR score, object coverage, hallucination rate, and alignment with human cognition.
\textbf{2) MMHalBench} \cite{sun2023llava_rlhf}: A question-answering benchmark with 96 images across 12 object categories. Following the official protocol, we use GPT-4 to rate responses from zero to six, calculating hallucination rate by the proportion of responses rated below three.
\textbf{3) Object HalBench} \cite{rohrbach2018objecthal}: A widely used benchmark for assessing object hallucination. We evaluate across 300 instances using the Yu et al. \cite{yu2024rlhfv} codebase, reporting hallucination rates at both response (CHAIRs) and object levels (CHAIRi).\footnote{It is noted that some studies report results based on 50 instances, we exclude these outcomes to prevent any potential confusion.}
\textbf{4) POPE} \cite{li2023pope}: A yes/no question-answering benchmark for object hallucination evaluation. We report accuracy and precision on its Adversarial set, consisting of 3000 cases.
\vspace{-3mm}
\paragraph{Baseline Algorithms.}
We mainly compare OPA-DPO with algorithms based on RLHF/RLAIF. As mentioned in Chapter \ref{sec:intro}, most algorithms, such as HALVA \cite{sarkar2024halva}, POVID \cite{zhou2024povid}, RLHF-V \cite{yu2024rlhfv}, HA-DPO \cite{zhao2023hadpo}, HSA-DPO \cite{xiao2024hsadpo}, RLAIF-V \cite{yu2024rlaif}, and mDPO \cite{wang2024mdpo}, prefer to use DPO,
while LLaVA-RLHF \cite{sun2023llava_rlhf} choose to use PPO.
\vspace{-3mm}
\paragraph{Implementation Details.}
For both the 7B and 13B models, we start with OPA training (LoRA-SFT) for 2 epochs, using a cosine learning rate schedule beginning at 2e-5. We set the batch size to 128, with a LoRA rank of 256 and alpha of 512. Following this, we perform OPA-DPO training on the SFT-tuned LoRA module for 4 more epochs, using a batch size of 32 and a cosine learning rate starting at 1e-6.
In our equations, we set $\beta=0.1$ in Eqs.~\eqref{eq7}\eqref{eq8}\eqref{eq9}, and $\gamma_1=0.2, \gamma_2=1.0$ in Eq.~\eqref{eq10}. For the distorted images $\mathbf{m'}$ in Eq.~\eqref{eq8}, we randomly mask 30\% of pixels. 
For the anchors in Eq.~\eqref{eq9}, we set $\delta=0$. 
See Appendix for pseudo code and more detailed settings.

\begin{figure}[t]
  \centering
   \includegraphics[width=0.99\linewidth]{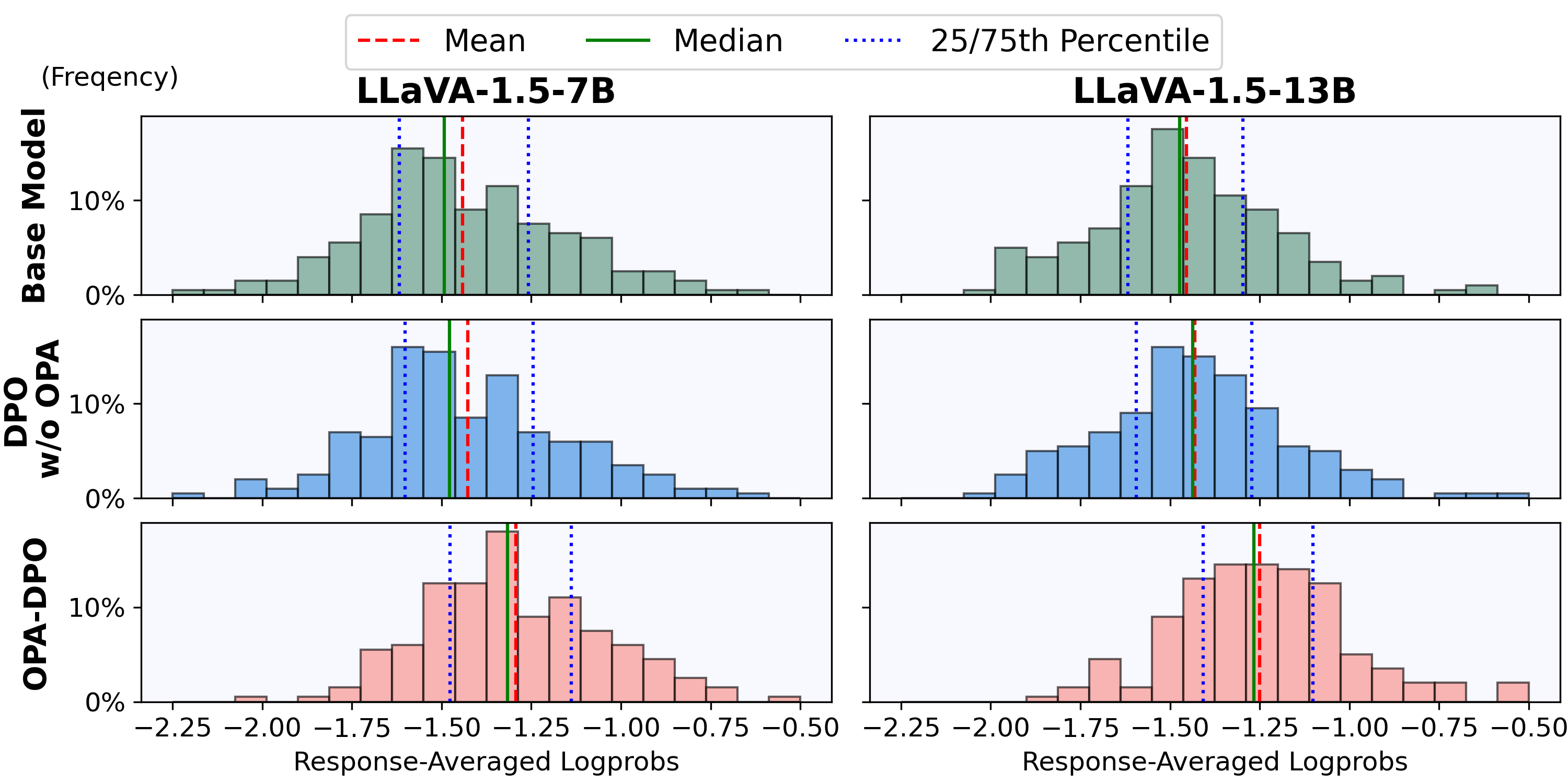}
   \vspace{-2mm}
   \caption{Distribution of response-averaged log probabilities for 200 significantly revised responses across different models.}\label{fig4_distribution}
   \vspace{-4mm}
\end{figure}

\subsection{Policy Distribution over Revised Responses}
To demonstrate that off-policy preferred responses are not effectively learned through DPO, we visualize the response-averaged log probabilities of tokens, denoted as $\frac{1}{L}\sum_i^L \log \pi(\mathbf{y}_i | \mathbf{x}, \mathbf{m}, \mathbf{y}_{<i})$, across 200 significantly revised responses from GPT-4V datasets, as shown in Figure~\ref{fig4_distribution}. 
The distribution shows negligible change after DPO training without OPA, whereas a significant increase is observed with our proposed OPA-DPO.


\begin{figure}[b]
  \centering
  \vspace{-4mm}
   \includegraphics[width=0.95\linewidth]{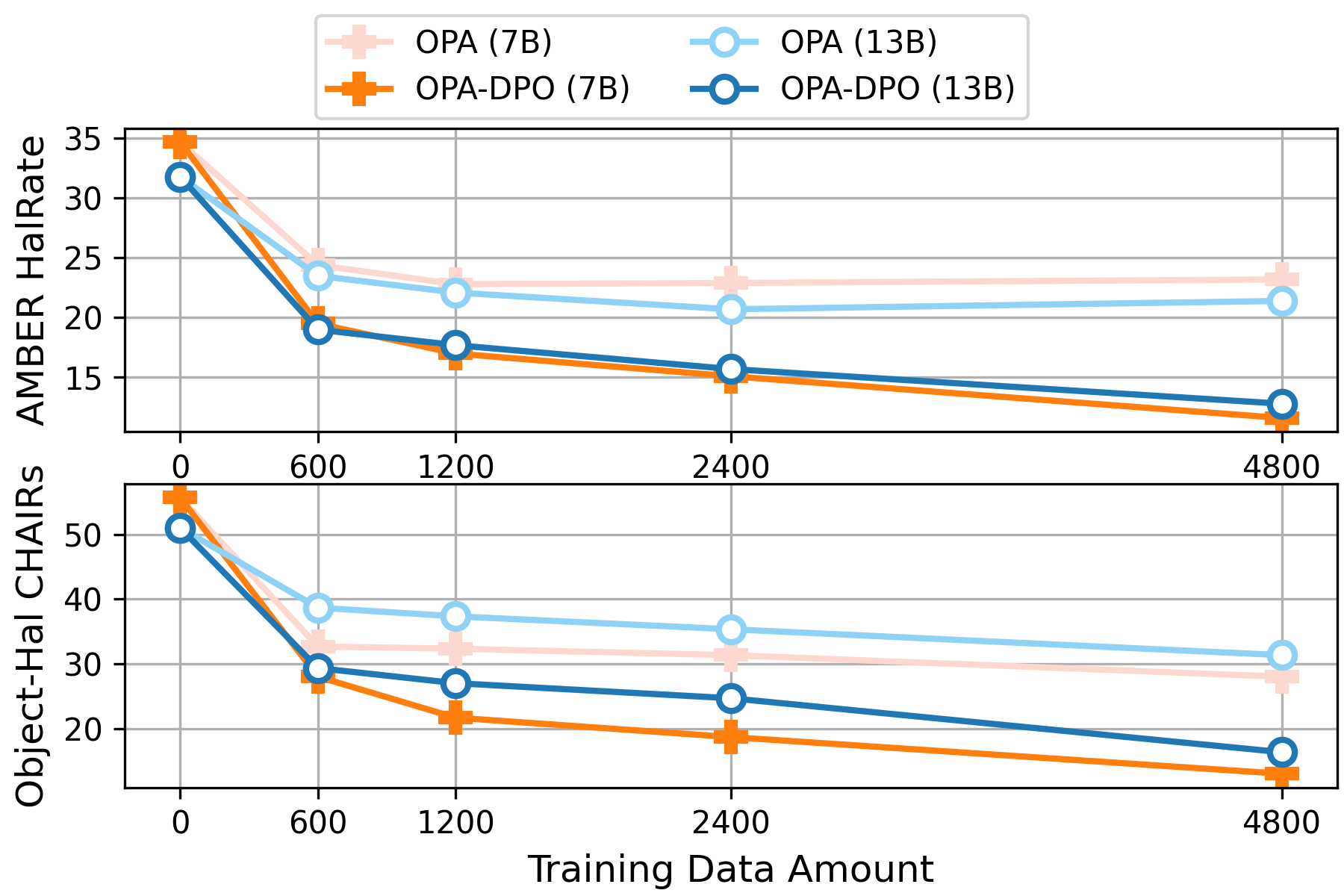}
   \vspace{-2mm}
   \caption{Impact of data amount on hallucination-rate metrics.}
   \vspace{-4mm}
   \label{fig5_data_scale}
\end{figure}

In order to emphasize the constraint arises from reverse KL-divergence, we measure the average KL-divergence {\small $\frac{1}{L}\sum_i^L \mathbb{D}_\mathrm{KL} [\pi_P(\cdot | \mathbf{x}, \mathbf{m}, \mathbf{y}_{<i}) \Vert \pi_Q(\cdot | \mathbf{x}, \mathbf{m}, \mathbf{y}_{<i})]$} and maximum KL-divergence {\small $\max_i \mathbb{D}_\mathrm{KL} [\pi_P(\cdot | \mathbf{x}, \mathbf{m}, \mathbf{y}_{<i}) \Vert \pi_Q(\cdot | \mathbf{x}, \mathbf{m}, \mathbf{y}_{<i})]$} between different policies, averaging the results over the same 200 samples in Table~\ref{table1_kl}.
We observe that the reverse KL shows minor change after DPO ({\small $\mathbb{D}_\mathrm{KL}[\pi_\mathrm{DPO}\Vert\pi_{base}]$} and {\small $\mathbb{D}_\mathrm{KL}[\pi^\mathrm{OPA}_\mathrm{DPO}\Vert\pi_\mathrm{OPA}]$});
however, the divergence gap between {\footnotesize $\pi^\mathrm{OPA}_\mathrm{DPO}$} and {\footnotesize $\pi_{base}$} is nearly an order of magnitude larger, indicating that naive DPO is insufficient to bridge this gap.

\begin{table}[t]
\caption{Comparison of average and maximum KL-divergence between different policies. Results are averaged over the same 200 significantly revised responses as in Figure~\ref{fig4_distribution}.}
\label{table1_kl}
\vspace{-2mm}
\centering
\footnotesize
\begin{tabular}{lllllll}
\hline \addlinespace[0.5mm]
\multicolumn{2}{l}{\multirow{2}{*}{$\mathbb{D}_\mathrm{KL}[P \, \Vert \, Q]$ $\qquad$}} & $P$ & $\pi_\mathrm{DPO}$ & $\pi^\mathrm{OPA}_\mathrm{DPO}$ & $\pi^\mathrm{OPA}_\mathrm{DPO}$ & $\pi_\mathrm{OPA}$ \\ \addlinespace[0.5mm] \cline{3-7} \addlinespace[0.5mm]
\multicolumn{2}{l}{} & $Q$ & $\pi_{base}$ & $\pi_\mathrm{base}$ & $\pi_\mathrm{OPA}$ & $\pi_\mathrm{base}$ \\ \addlinespace[0.5mm] \hline 
\multirow{2}{*}{7B} & \multicolumn{2}{l|}{$\mathrm{mean}-\mathrm{mean}$} & 0.039 & 0.371 & 0.025 & 0.276 \\
 & \multicolumn{2}{l|}{$\mathrm{max}-\mathrm{mean}$} & 0.396 & 2.288 & 0.161 & 1.839 \\ \hline
\multirow{2}{*}{13B} & \multicolumn{2}{l|}{$\mathrm{mean}-\mathrm{mean}$} & 0.044 & 0.261 & 0.036 & 0.174 \\
 & \multicolumn{2}{l|}{$\mathrm{max}-\mathrm{mean}$} & 0.421 & 1.944 & 0.349 & 1.364 \\ \hline
\end{tabular}
\vspace{-3mm}
\end{table}

\subsection{Benchmark Evaluation Results}\label{benchmark_main_results}
\begin{table*}[t]
\caption{Comparison of RLAIF/RLHF-based algorithms for enhancing LVLMs across various benchmarks. For baseline algorithms with available official checkpoints, we retest the models, and these results are marked with $\S$. For algorithms without official checkpoints, results are sourced from the respective papers: $\dagger$ denotes results from \cite{xiao2024hsadpo}, $\ddagger$ from \cite{sarkar2024halva}, and $\star$ from \cite{wang2024mdpo}. To ensure a fair comparison, greedy sampling is used in all evaluations to avoid potential randomness. The best result for each metric within each group is highlighted in \textbf{bold}.}\label{table2_main_result}
\vspace{-2mm}
\resizebox{\textwidth}{36mm}{
\begin{tabular}{lcc|cccccccccc}
\hline
\rowcolor[HTML]{EFEFEF} 
\multicolumn{1}{c}{\cellcolor[HTML]{EFEFEF}{\color[HTML]{000000} }} & \cellcolor[HTML]{EFEFEF}{\color[HTML]{000000} } & \cellcolor[HTML]{EFEFEF}{\color[HTML]{000000} } & \multicolumn{4}{c|}{\cellcolor[HTML]{EFEFEF}{\color[HTML]{000000} \textbf{\begin{tabular}[c]{@{}c@{}}AMBER\\ (1004)\end{tabular}}}} & \multicolumn{2}{c|}{\cellcolor[HTML]{EFEFEF}{\color[HTML]{000000} \textbf{\begin{tabular}[c]{@{}c@{}}MMHal-Bench\\ (96)\end{tabular}}}} & \multicolumn{2}{c|}{\cellcolor[HTML]{EFEFEF}{\color[HTML]{000000} \textbf{\begin{tabular}[c]{@{}c@{}}Object Hal \\ (300)\end{tabular}}}} & \multicolumn{2}{c}{\cellcolor[HTML]{EFEFEF}{\color[HTML]{000000} \textbf{\begin{tabular}[c]{@{}c@{}}POPE Adversarial\\ (3000)\end{tabular}}}} \\ \cline{4-13} 
\rowcolor[HTML]{EFEFEF} 
\multicolumn{1}{c}{\multirow{-2}{*}{\cellcolor[HTML]{EFEFEF}{\color[HTML]{000000} \textbf{\begin{tabular}[c]{@{}c@{}}Algorithm\\ \,\end{tabular}}}}} & \multirow{-2}{*}{\cellcolor[HTML]{EFEFEF}{\color[HTML]{000000} \textbf{\begin{tabular}[c]{@{}c@{}}Data Size\\ \,\end{tabular}}}} & \multirow{-2}{*}{\cellcolor[HTML]{EFEFEF}{\color[HTML]{000000} \textbf{\begin{tabular}[c]{@{}c@{}}Feedback\\ \,\end{tabular}}}} & {\color[HTML]{000000} \textbf{CHAIR↓}} & {\color[HTML]{000000} \textbf{Cover↑}} & {\color[HTML]{000000} \textbf{HalRate↓}} & \multicolumn{1}{c|}{\cellcolor[HTML]{EFEFEF}{\color[HTML]{000000} \textbf{Cog↓}}} & {\color[HTML]{000000} \textbf{Score↑}} & \multicolumn{1}{c|}{\cellcolor[HTML]{EFEFEF}{\color[HTML]{000000} \textbf{HalRate↓}}} & {\color[HTML]{000000} \textbf{CHAIRs↓}} & \multicolumn{1}{c|}{\cellcolor[HTML]{EFEFEF}{\color[HTML]{000000} \textbf{CHAIRi↓}}} & {\color[HTML]{000000} \textbf{Acc.↑}} & {\color[HTML]{000000} \textbf{Pre.↑}} \\ \hline \addlinespace
\multicolumn{3}{l|}{\textbf{Qwen-VL-Chat -34B \cite{bai2023qwenvl}$^\dagger$}} & 6.6 & 53.2 & 31.0 & 2.9 & 2.89 & 0.43 & 36 & 21.3 & - & - \\
+Silkie \cite{li2023silkie}$^\dagger$ & 80k & GPT-4V & 5.4 & 55.8 & 29.0 & 2.0 & 3.01 & 0.41 & 25.3 & 13.9 & - & - \\ \addlinespace \hline \addlinespace
\multicolumn{3}{l|}{\textbf{LLaVA-Instruct-1.5-7B \cite{liu2024llava, liu2024llava2}$^\S$}} & 7.7 & 51.6 & 34.7 & 4.2 & 2.01 & 0.61 & 55.67 & 15.96 & \textbf{84.93\%} & 89.10\% \\
+LLaVA-RLHF \cite{sun2023llava_rlhf}$^\S$ & 122k & Self-Reward & 9.7 & 53.2 & 46.6 & 5.3 & 1.88 & 0.71 & 58.00 & 15.61 & 80.00\% & 87.19\% \\
+HALVA \cite{sarkar2024halva}$^\ddagger$ & 21.5k & GPT-4V & 6.6 & \textbf{53.0} & 32.2 & 3.4 & 2.25 & 0.54 & 41.40 & 11.70 & - & - \\
+mDPO \cite{wang2024mdpo}$^\star$ & 10k & GPT-4V & 4.4 & 52.4 & 24.5 & 2.4 & 2.39 & 0.54 & 35.70 & 9.80 & - & 95.36\% \\
+HA-DPO \cite{zhao2023hadpo}$^\S$ & 6k & GPT4 & 7.8 & 52.1 & 35.6 & 4.2 & 1.89 & 0.65 & 54.00 & 14.45 & 84.90\% & 90.42\% \\
+POVID \cite{zhou2024povid}$^\S$ & 17k & GPT-4V & 7.4 & 51.3 & 34.3 & 3.9 & 2.08 & 0.60 & 50.67 & 15.28 & 84.77\% & 89.01\% \\
+RLAIF-V \cite{yu2024rlaif}$^\S$ & 16k & LLaVA-Next & 3.0 & 50.4 & 16.2 & 1.0 & \textbf{3.00} & \textbf{0.38} & 16.00 & \textbf{3.70} & 81.57\% & 94.97\% \\
\rowcolor[HTML]{ECF4FF} 
\textbf{+OPA (ours)} & 4.8k & GPT-4V & 5.6 & 52.8 & 23.2 & 2.3 & 2.41 & 0.52 & 28.00 & 9.48 & 82.53\% & 95.36\% \\
\rowcolor[HTML]{ECF4FF} 
\textbf{+OPA-DPO (ours)} & 4.8k & GPT-4V & \textbf{2.2} & 47.9 & \textbf{11.6} & \textbf{0.9} & 2.83 & 0.45 & \textbf{13.00} & 4.25 & 82.60\% & \textbf{95.61\%} \\ \addlinespace \hline \addlinespace
\multicolumn{3}{l|}{\textbf{LLaVA-Instruct-1.5-13B \cite{liu2024llava, liu2024llava2}$^\S$}} & 6.8 & 51.9 & 31.8 & 3.3 & 2.48 & 0.52 & 51.00 & 13.71 & \textbf{85.50\%} & 90.31\% \\
+LLaVA-RLHF \cite{sun2023llava_rlhf}$^\S$ & 122k & Self-Reward & 7.7 & 52.3 & 38.6 & 4.0 & 2.27 & 0.64 & 44.67 & 11.83 & 82.47\% & 90.25\% \\
+RLHF-V (HD) \cite{yu2024rlhfv}$^\dagger$ & 1.4k & Human & 6.3 & 46.1 & 25.1 & 2.1 & 2.81 & 0.49 & - & - & - & - \\
+HSA-DPO \cite{xiao2024hsadpo}$^\dagger$ & 8k & GPT-4/4V & \textbf{2.1} & 47.3 & 13.4 & 1.2 & 2.61 & 0.48 & - & - & 84.00\% & 80.20\% \\
+HALVA \cite{sarkar2024halva}$^\ddagger$ & 21.5k & GPT-4V & 6.4 & 52.6 & 30.4 & 3.2 & 2.58 & 0.45 & 45.40 & 12.80 & - & - \\
\rowcolor[HTML]{ECF4FF} 
\textbf{+OPA (ours)} & 4.8k & GPT-4V & 5.2 & \textbf{54.1} & 21.4 & 2.2 & 2.75 & 0.45 & 31.33 & 8.88 & 83.60\% & 96.24\% \\
\rowcolor[HTML]{ECF4FF} 
\textbf{+OPA-DPO (ours)} & 4.8k & GPT4V & 2.4 & 48.3 & \textbf{12.8} & \textbf{0.9} & \textbf{3.07} & \textbf{0.39} & \textbf{16.33} & \textbf{5.48} & 82.63\% & \textbf{96.31\%} \\ \addlinespace \hline
\end{tabular}
}
\vspace{-2mm}
\end{table*}

The experimental results across various benchmarks are presented in Table~\ref{table2_main_result}. For LLaVA-Instruct-1.5-7B, our OPA-DPO achieves SOTA performance in 50\% of the hallucination metrics, which increases to 70\% for the LLaVA-Instruct-1.5-13B. OPA-DPO particularly excels in metrics that measure the occurrence of hallucinations, such as CHAIR and HalRate. 
However, the enhancement leads to a slight compromise in coverage-related metrics (Cover). In the yes-or-no benchmark (POPE), while precision significantly improves, accuracy remains the same due to the model's tendency to provide fewer 'yes' answers. This results in higher accuracy for positive samples but lower accuracy for negative ones. All indicators suggest that the OPA-DPO trained models tend to adopt a slightly conservative strategy, avoiding uncertain assertions. This strategy significantly enhances the credibility of responses but may omit some ambiguous details, which necessitate a trade-off. 

To demonstrate the scalability of OPA-DPO, we present its performance under varying amounts of training data as in Figure~\ref{fig5_data_scale}. Even with 600 data only, OPA-DPO surpasses the majority of baseline algorithms in metrics related to hallucinations. Notably, increasing the data volume does not lead to significant performance improvements in $\pi_\mathrm{OPA}$, the policy after LoRA-SFT. However, the performance enhancement of OPA-DPO with increased data is quite remarkable.

\subsection{Ablation Studies}\label{sec_ablation}
We emphasize that each component of our OPA-DPO, as detailed in Chapter~\ref{sec_OPADPO}, is important. The OPA operation, specifically the LoRA-SFT on GT responses and the GPT-4V revised response, is the most critical element.

\begin{table}[h]
\vspace{-2mm}
\caption{Ablation studies on On-Policy Alignment operation.}\label{table3_ablation_OPA}
\vspace{-2mm}
\scriptsize
\setlength{\tabcolsep}{1pt}
\begin{tabular}{cllcccccc}
\hline
\rowcolor[HTML]{EFEFEF} 
\cellcolor[HTML]{EFEFEF} & \multicolumn{1}{c}{\cellcolor[HTML]{EFEFEF}} & \multicolumn{1}{c|}{\cellcolor[HTML]{EFEFEF}} & \multicolumn{4}{c|}{\cellcolor[HTML]{EFEFEF}\textbf{AMBER}} & \multicolumn{2}{c}{\cellcolor[HTML]{EFEFEF}\textbf{Object Hal}} \\ \cline{4-9} 
\rowcolor[HTML]{EFEFEF} 
\multirow{-2}{*}{\cellcolor[HTML]{EFEFEF}\textbf{\begin{tabular}[c]{@{}c@{}}Model\\size\end{tabular}}} & \multicolumn{1}{c}{\multirow{-2}{*}{\cellcolor[HTML]{EFEFEF}\textbf{\begin{tabular}[c]{@{}c@{}}Data\\size\end{tabular}}}} & \multicolumn{1}{c|}{\multirow{-2}{*}{\cellcolor[HTML]{EFEFEF}\textbf{Algo.}}} & \textbf{CHAIR↓} & \textbf{Cover↑} & \textbf{HalRate↓} & \multicolumn{1}{c|}{\cellcolor[HTML]{EFEFEF}\textbf{Cog↓}} & \textbf{CHAIRs↓} & \textbf{CHAIRi↓} \\ \hline \addlinespace[0.5mm]
 &  & \cellcolor[HTML]{ECF4FF}\textbf{w OPA} & \cellcolor[HTML]{ECF4FF}2.2 & \cellcolor[HTML]{ECF4FF}47.9 & \cellcolor[HTML]{ECF4FF}11.6 & \cellcolor[HTML]{ECF4FF}0.9 & \cellcolor[HTML]{ECF4FF}13.00 & \cellcolor[HTML]{ECF4FF}4.25 \\
 & \multirow{-2}{*}{\textbf{4.8k}} & \textbf{w/o OPA} & 3.8 & 48.0 & 22.6 & 2.2 & 23.00 & 7.64 \\ \addlinespace[0.5mm] \cline{2-9} \addlinespace[0.5mm]
 &  & \cellcolor[HTML]{ECF4FF}\textbf{w OPA} & \cellcolor[HTML]{ECF4FF}3.3 & \cellcolor[HTML]{ECF4FF}47.8 & \cellcolor[HTML]{ECF4FF}15.1 & \cellcolor[HTML]{ECF4FF}1.3 & \cellcolor[HTML]{ECF4FF}18.67 & \cellcolor[HTML]{ECF4FF}5.63 \\
\multirow{-4}{*}{\textbf{7B}} & \multirow{-2}{*}{\textbf{2.4k}} & \textbf{w/o OPA} & 4.6 & 48.6 & 26.8 & 1.8 & 34.67 & 9.81 \\ \addlinespace[0.5mm] \hline \addlinespace[0.5mm]
 &  & \cellcolor[HTML]{ECF4FF}\textbf{w OPA} & \cellcolor[HTML]{ECF4FF}2.4 & \cellcolor[HTML]{ECF4FF}48.3 & \cellcolor[HTML]{ECF4FF}12.8 & \cellcolor[HTML]{ECF4FF}0.9 & \cellcolor[HTML]{ECF4FF}16.33 & \cellcolor[HTML]{ECF4FF}5.48 \\
 & \multirow{-2}{*}{\textbf{4.8k}} & \textbf{w/o OPA} & 5.7 & 50.4 & 27.5 & 2.7 & 32.67 & 9.45 \\ \addlinespace[0.5mm] \cline{2-9} \addlinespace[0.5mm]
 &  & \cellcolor[HTML]{ECF4FF}\textbf{w OPA} & \cellcolor[HTML]{ECF4FF}4.1 & \cellcolor[HTML]{ECF4FF}49.8 & \cellcolor[HTML]{ECF4FF}15.7 & \cellcolor[HTML]{ECF4FF}1.4 & \cellcolor[HTML]{ECF4FF}24.67 & \cellcolor[HTML]{ECF4FF}7.38 \\
\multirow{-4}{*}{\textbf{13B}} & \multirow{-2}{*}{\textbf{2.4k}} & \textbf{w/o OPA} & 5.2 & 49.7 & 25.6 & 2.7 & 38.33 & 11.98 \\ \addlinespace[0.5mm] \hline
\end{tabular}
\vspace{-2mm}
\end{table}

To highlight the significance of our proposed On-Policy Alignment framework in training DPO, we conduct ablation studies on the OPA operation, as illustrated in Table~\ref{table3_ablation_OPA}. 
The results indicate that the performance of the trained policy without OPA is nearly identical to that of RLHF-V and mDPO, neither of which account for on-policy data.
However, integrating the OPA operation results in a nearly 50\% reduction in the AMBER HalRate and Object-hal CHAIRs metrics compared to the policy trained without OPA.

\begin{table}[h]
\caption{Ablation studies on the Image Focus mechanism (IF), Anchored preference (Anc), and the hallucination-weighted (hw) and image-weighted (iw) policy updating. The metric ``repeat" indicates the frequency of generating sentence- or phrase-level repetitions without an EOS token across 1004 AMBER samples.}\label{table4_ablation_component}
\vspace{-2mm}
\scriptsize
\resizebox{0.47\textwidth}{17mm}{
\setlength{\tabcolsep}{1pt}
\begin{tabular}{cl|ccccc|cc}
\hline
\rowcolor[HTML]{EFEFEF} 
\cellcolor[HTML]{EFEFEF} & \cellcolor[HTML]{EFEFEF} & \multicolumn{5}{c|}{\cellcolor[HTML]{EFEFEF}\textbf{AMBER}} & \multicolumn{2}{c}{\cellcolor[HTML]{EFEFEF}\textbf{Object Hal}} \\ \cline{3-9} 
\rowcolor[HTML]{EFEFEF} 
\multirow{-2}{*}{\cellcolor[HTML]{EFEFEF}\textbf{\begin{tabular}[c]{@{}c@{}}Model\\size\end{tabular}}} & \multirow{-2}{*}{\cellcolor[HTML]{EFEFEF}\textbf{Ablation}} & \textbf{CHAIR↓} & \textbf{Cover↑} & \textbf{HalRate↓} & \textbf{Cog↓} & \textbf{repeat} & \textbf{CHAIRs↓} & \textbf{CHAIRi↓} \\ \hline \addlinespace[0.5mm]
 & \cellcolor[HTML]{ECF4FF}\textbf{OPA-DPO} & \cellcolor[HTML]{ECF4FF}2.2 & \cellcolor[HTML]{ECF4FF}47.9 & \cellcolor[HTML]{ECF4FF}11.6 & \cellcolor[HTML]{ECF4FF}0.9 & \cellcolor[HTML]{ECF4FF}0.6\% & \cellcolor[HTML]{ECF4FF}13.00 & \cellcolor[HTML]{ECF4FF}4.25 \\
 & \textbf{w/o IF} & 5.1 & 50.7 & 15.4 & 1.1 & 15.7\% & 14.67 & 9.87 \\
 & \textbf{w/o Anc} & 2.3 & 45.6 & 13.2 & 1.0 & 6.7\% & 14.33 & 4.21 \\
 & \textbf{w/o IF\&Anc} & 4.2 & 50.4 & 16.2 & 1.2 & 15.1\% & 13.00 & 9.63 \\
\multirow{-5}{*}{\textbf{7B}} & \textbf{w/o hw\&iw} & 2.4 & 46.2 & 12.6 & 0.9 & 0.4\% & 17.00 & 4.68 \\ \addlinespace[0.5mm] \hline \addlinespace[0.5mm]
 & \cellcolor[HTML]{ECF4FF}\textbf{OPA-DPO} & \cellcolor[HTML]{ECF4FF}2.4 & \cellcolor[HTML]{ECF4FF}48.3 & \cellcolor[HTML]{ECF4FF}12.8 & \cellcolor[HTML]{ECF4FF}0.9 & \cellcolor[HTML]{ECF4FF}0.8\% & \cellcolor[HTML]{ECF4FF}16.33 & \cellcolor[HTML]{ECF4FF}5.48 \\
 & \textbf{w/o IF} & 3.2 & 53.1 & 16.9 & 1.3 & 17.1\% & 21.33 & 9.82 \\
 & \textbf{w/o Anc} & 2.4 & 48.5 & 13.5 & 0.9 & 4.6\% & 17.33 & 5.38 \\
 & \textbf{w/o IF\&Anc} & 3.5 & 52.9 & 16.7 & 1.2 & 15.9\% & 21.33 & 12.36 \\
\multirow{-5}{*}{\textbf{13B}} & \textbf{w/o hw\&iw} & 2.8 & 48.8 & 14.6 & 1.1 & 0.9\% & 18.33 & 6.02 \\ \addlinespace[0.5mm] \hline
\end{tabular}
}
\vspace{-2mm}
\end{table}

In addition, we present ablation studies on the three components of our OPA-DPO training loss as described in Chapter~\ref{OPA_DPO}. The evaluation results, shown in Table~\ref{table4_ablation_component}, indicate that each term in Eq.\eqref{eq10}, as well as the hallucination-weighted/image-weighted policy updates in Eqs.\eqref{eq8} and \eqref{eq9}, plays a crucial role in reducing hallucination. Notably, in the absence of the Image Focus mechanism (IF) or Anchored Preference (Anc), the policy tends to repeat its last sentence or words and fails to generate an EOS token when using greedy sampling. This phenomenon is particularly pronounced in long-form QA tasks, such as the AMBER generation task. However, when all three components are employed together, the repetition issue is resolved.

\subsection{Case Study}\label{sec5_5case_study}
To provide an intuitive understanding of our OPA-DPO, we present a qualitative example in Figure~\ref{fig6_case}. The initial model's generation contains numerous hallucinations and flawed reasoning. This issue persists after training naive DPO without OPA. However, after implementing OPA on 4.8k samples, the hallucinations are nearly eliminated, though some minor instances remain. Subsequent OPA-DPO completely resolves these issues, albeit at the cost of omitting some details present in the original description, which aligns with our discussion in Chapter~\ref{benchmark_main_results}.

\section{Related Works}

\definecolor{hl_blue}{HTML}{70F3FF}
\definecolor{hl_yellow}{HTML}{FFFF00}
\begin{figure}[t]
  \centering
   \includegraphics[width=0.99\linewidth]{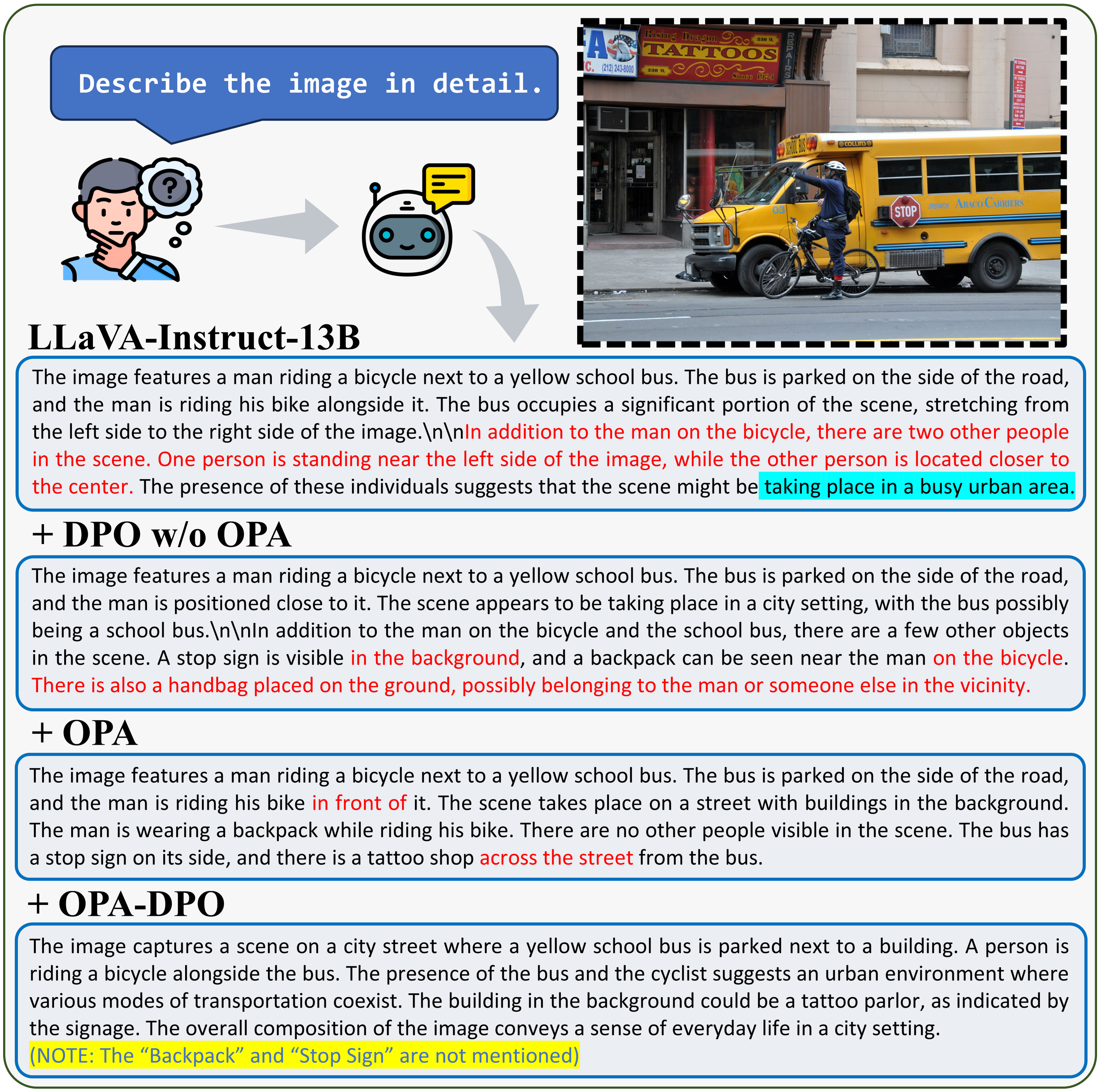}
   \vspace{-2mm}
   \caption{Qualitative example of responses from different models with the same prompt and image. Hallucinated parts are marked in \textcolor{red}{red}, flawed reasoning is {\sethlcolor{hl_blue}\hl{highlighted in blue}}, and missing details are {\sethlcolor{hl_yellow}\hl{highlighted in yellow}}. This example illustrates a common case; additional examples can be found in the Appendix.}\label{fig6_case}
   \vspace{-5mm}
\end{figure}

\paragraph{RLHF.}
As a fundamental technique driving the advancements of LLMs and LVLMs in recent years, RLHF \cite{christiano2017deep, ziegler2019fine, ouyang2022training} has been demonstrated to be effective in aligning fine-tuned large models with human preferences. By leveraging vast amounts of human preference data and RL methodologies, numerous language models have benefited from this approach and have been widely adopted. Notable examples include GPT \cite{radford2018gpt, radford2019gpt, brown2020gpt, achiam2023gpt}, LLaMA \cite{dubey2024llama, touvron2023llama2, touvron2023llama}, Qwen \cite{bai2023qwen, chu2023qwen, yang2024qwen2}, Gemini \cite{team2023gemini, team2024gemini}, and Claude \cite{TheC3}.
PPO \cite{schulman2017ppo} is the original RL algorithm used in RLHF. While stable, its reliance on a dependable reward model and numerous hyper-parameters has led to the exploration of alternatives. DPO \cite{rafailov2024dpo} has attracted attention for its strong performance and removal of the need for a separate reward model. However, it has not yet matched PPO's performance \cite{xu2024dpo_ppo}, motivating efforts to close this gap. Methods such as ORPO \cite{hong2024orpo}, CPO \cite{xu2024cpo}, TPO \cite{saeidi2024tpo}, and SimPO \cite{meng2024simpo} aim to better align models with preference data by removing reference policy constraints. Nevertheless, these approaches lack comprehensive validation across various datasets and modalities. More pertinent to our work, iterative DPO \cite{pang2024iterdpo} and SPPO \cite{wu2024sppo} address off-policy issues by sampling preferred responses on-policy. This manner is adopted by RLAIF-V \cite{yu2024rlaif}, but it faces challenges with low efficiency in addressing persistent hallucinations in multimodal contexts.

\vspace{-4mm}
\paragraph{Hallucination for LVLMs.}
Hallucination reduction in LVLMs has garnered significant attention as a major misalignment issue \cite{rani2024visual, bai2024hal_survey, liu2024survey}. We categorize the methodologies into two classes. The first class, termed RL-free, primarily investigates the decoding mechanisms of translating visual information into language output, with some studies focusing on attention patterns \cite{gong2024damro, huang2024opera, yuan2024helpd} and others examining distribution shifts when image information is distorted \cite{wang2024icd, leng2024vcd}. Additionally, some research explores the intriguing effects of special tokens on hallucinations, such as 'EOS' \cite{yue2024eos} and '$\backslash$n' \cite{han2024n}. 
\setlength{\intextsep}{5pt} 
\setlength{\columnsep}{5pt} 
\begin{wraptable}{r}{5.3cm}
\caption{Comparison of algorithms utilizing DPO to address hallucinations.}\label{table5_summerize}
\vspace{-2mm}
\footnotesize
\resizebox{0.3\textwidth}{14mm}{
\begin{tabular}{lccc}
\hline
\rowcolor[HTML]{EFEFEF} 
\textbf{Algorithm} & \textbf{\begin{tabular}[c]{@{}c@{}}Expert\\ Correction\end{tabular}} & \textbf{\begin{tabular}[c]{@{}c@{}}On-policy \\ Data\end{tabular}} & \textbf{\begin{tabular}[c]{@{}c@{}}Image \\ Focus\end{tabular}} \\ \hline \addlinespace[0.5mm]
HALVA\cite{sarkar2024halva} & \textcolor{red}{\ding{55}} & \textcolor{red}{\ding{55}} & \textcolor{red}{\ding{55}} \\
POVID\cite{zhou2024povid} & \textcolor{red}{\ding{55}} & \textcolor{red}{\ding{55}} & \textcolor{Green}{\ding{51}} \\
RLHF-V\cite{yu2024rlhfv} & \textcolor{Green}{\ding{51}} & \textcolor{red}{\ding{55}} & \textcolor{red}{\ding{55}} \\
HA-DPO\cite{zhao2023hadpo} & \textcolor{Green}{\ding{51}} & \textcolor{red}{\ding{55}} & \textcolor{red}{\ding{55}} \\
HSA-DPO\cite{xiao2024hsadpo} & \textcolor{Green}{\ding{51}} & \textcolor{red}{\ding{55}} & \textcolor{red}{\ding{55}} \\
RLAIF-V\cite{yu2024rlaif} & \textcolor{red}{\ding{55}} & \textcolor{Green}{\ding{51}} & \textcolor{red}{\ding{55}} \\
mDPO\cite{wang2024mdpo} & \textcolor{red}{\ding{55}} & \textcolor{red}{\ding{55}} & \textcolor{Green}{\ding{51}} \\
\rowcolor[HTML]{ECF4FF} 
\textbf{OPA-DPO} & \textcolor{Green}{\ding{51}} & \textcolor{Green}{\ding{51}} & \textcolor{Green}{\ding{51}} \\ \hline
\end{tabular}
}
\end{wraptable}
The second class, RL-based, employs the RLHF framework to gather feedback from humans or AI systems with superhuman capabilities. Compared with RL-freed methods, RL-based methods generally demonstrate superior results on benchmarks designed to assess the reduction of hallucination. Within this class, only a few studies elect PPO \cite{sun2023llava_rlhf}, while the majority, like our work, choose DPO \cite{sarkar2024halva, zhou2024povid, yu2024rlhfv, zhao2023hadpo, xiao2024hsadpo, yu2024rlaif, wang2024mdpo, li2023silkie}. Given the inherent vulnerabilities associated with the naive adoption of DPO in LVLMs, we summarize the characteristics of various algorithms across three dimensions: expert correction, on-policy data, and image-focus, as presented in Table~\ref{table5_summerize}. Our OPA-DPO is the only algorithm that considers all three aspects, thereby achieving SOTA performance across multiple metrics.
\vspace{-2mm}
\section{Conclusions}
\vspace{-2mm}
In conclusion, our study uncovers a crucial characteristic of DPO: heavy reliance on on-policy data. By examining dataset distribution, we identify and systematically summarize the inherent flaws in existing DPO-based algorithms for addressing hallucination issues. 
To address the shortcomings, we introduce On-Policy Alignment (OPA)-DPO, a framework that integrates the strengths of various approaches. 
OPA-DPO leverages expert feedback to correct hallucinated responses and ensures alignment of both the original and expert-revised responses on-policy. Remarkably, with only 4.8k training samples, OPA-DPO improved LLaVA-1.5-7B and LLaVA-1.5-13B achieve SOTA performance on over half hallucination-related benchmarks, surpassing other DPO-based algorithms in mitigating hallucination problems, which generally requires over 10k data.
\section*{Acknowledgments}
{
This work was supported in part by the General Research Fund (GRF) project 14200720 of the Hong Kong University Grants Committee and the National Natural Science Foundation of China (NSFC) Project 62073273. This work was also partially supported by Microsoft Research. }

{
\clearpage
\maketitlesupplementary
\appendix

\noindent The Appendix is organized as follows:

\begin{itemize}
\item In Chapter~\ref{sec_implementation}, we offer a comprehensive description of our implementation details\footnote{Our implementation is available at \url{https://github.com/zhyang2226/OPA-DPO}. We also provide OPA-DPO trained LLaVA-v1.5-\href{https://huggingface.co/zhyang2226/opadpo-lora_llava-v1.5-7b}{7B} and \href{https://huggingface.co/zhyang2226/opadpo-lora_llava-v1.5-13b}{13B} models, along with the corresponding \href{https://drive.google.com/drive/folders/1Xmrb43zIbk3IzLLRQx65iBf7J4SHXa9j?usp=drive_link}{training dataset}.}, 
complementing the information presented in Chapter~\ref{sec_exp_details}. 
The training details and hyperparameter settings are reported in Chapter~\ref{app_detail}, while the GPT-4V prompt and corresponding examples are provided in Chapter~\ref{app_prompt}.

\item In Chapter~\ref{sec_add_exps}, we supply some additional experimental results. The helpfulness-related benchmark evaluations are conducted in Chapter~\ref{sec_add_llavabench}, and additional ablation studies on the hyperparameter choosing are presented in Chapter~\ref{sec_add_ablation}.

\item In Chapter~\ref{sec_qual_results}, we provide additional analytical examples to complement the example presented in Figure~\ref{fig6_case}.

\end{itemize}

\section{Implementation Details}
\label{sec_implementation}

\subsection{Training Details.}\label{app_detail}

Importantly, we emphasize that OPA-DPO does not depend on detailed hyperparameter tuning for different base models or training datasets. In our experiments, we apply OPA-DPO to two LVLMs with varying parameter sizes: LLAVA-v1.5-7B and LLAVA-v1.5-13B. We maintain consistent hyperparameter settings for OPA-DPO across different models and training datasets.

As shown in Figure~\ref{fig3_OPA-DPO}, the initial step in OPA-DPO involves instructing the model (slated for training) to generate responses $\mathbf{y}_\mathrm{Gen}$ based on pre-collected images $\mathbf{m}$ and prompts $\mathbf{x}$. Notably, we employ a combination of "top-k" and "top-p" sampling methods to select tokens with relatively high sampling probabilities according to the initial policy, thereby revealing the intrinsic hallucinations of the policy itself. For token sampling, we set $\mathrm{topk}=30$, $\mathrm{topp}=0.95$, and use a temperature of $1.0$.

\begin{algorithm}[t]
\caption{{OPA-DPO Training}}\label{alg_opadpo}
\begin{algorithmic}[1]
\scriptsize
\Statex \hskip-2.4em \textbf{Phase 1 Training: On-Policy Alignment}
\Require Initial policy $\pi_\theta$, datasets $\mathcal{D}=\{\mathbf{x}, \mathbf{m}, \mathbf{y}_\mathrm{GT}, \mathbf{y}_\mathrm{Rev}\}^N$
\For{SFT epochs}
\For{$\{\mathbf{x}, \mathbf{m}, \mathbf{y}_\mathrm{GT}, \mathbf{y}_\mathrm{Rev}\}^{M_1} \sim \mathcal{D}$}
\State Calculate loss in Eq.~\eqref{eq4} for $\pi_\theta(\mathbf{y}_\mathrm{GT} \vert \mathbf{x}, \mathbf{m})$ and $\pi_\theta(\mathbf{y}_\mathrm{Rev} \vert \mathbf{x}, \mathbf{m})$
\State Update $\pi_\theta$
\EndFor
\EndFor\\
\Return OPA policy $\pi_\mathrm{OPA} = \pi_\theta^\mathrm{final}$
\vspace{1mm}
\Statex \hskip-2.4em \textbf{Phase 2 Training: OPA-DPO}
\Require Initial policy $\pi_{\theta'}=\pi_\mathrm{OPA}$; hyperparameters $\beta, \gamma_1, \gamma_2, \delta$; datasets $\mathcal{D}=\{\mathbf{x}, \mathbf{m}, \mathbf{y}_\mathrm{Gen}, \mathbf{y}_\mathrm{GT}, \mathbf{y}_\mathrm{Rev}, S_\mathrm{hal}, S_\mathrm{img}\}^N$
\For{OPA-DPO epochs}
\For{$\{\mathbf{x}, \mathbf{m}, \mathbf{y}_\mathrm{Gen}, \mathbf{y}_\mathrm{GT}, \mathbf{y}_\mathrm{Rev}, S_\mathrm{hal}, S_\mathrm{img}\}^{M_2} \sim \mathcal{D}$}
\State Calculate loss in Eq.~\eqref{eq7} with $\{\mathbf{x}, \mathbf{m}, \mathbf{y}_\mathrm{Gen}, \mathbf{y}_\mathrm{GT}, \mathbf{y}_\mathrm{Rev}, S_\mathrm{hal}\}^{M_2}$
\State Produce distorted image $\mathbf{m}' = \mathbf{m} \odot \mathrm{pixel\_mask}$
\State Calculate loss in Eq.~\eqref{eq8} with $\{\mathbf{x}, \mathbf{m}, \mathbf{m}', \mathbf{y}_\mathrm{GT}, \mathbf{y}_\mathrm{Rev}, S_\mathrm{img}\}^{M_2}$
\State Calculate loss in Eq.~\eqref{eq9} with $\{\mathbf{x}, \mathbf{m}, \mathbf{y}_\mathrm{GT}, \mathbf{y}_\mathrm{Rev}\}^{M_2}$
\State Combine the losses as in Eq.~\eqref{eq10}
\State Update $\pi_{\theta'}$
\EndFor
\EndFor\\
\Return OPA-DPO policy $\pi_\mathrm{OPA}^\mathrm{DPO} = \pi_{\theta'}^\mathrm{final}$

\end{algorithmic}
\end{algorithm}

Following that, GPT-4V is tasked with identifying hallucinations by evaluating the generated responses at the sentence level. 
Each sentence in a response is assigned a hallucination severity score, $S_\mathrm{hal}$, on a scale from one to four, indicating the severity of any hallucination present. As we introduced in Eq.~\ref{eq7}, this score is incorporated into hallucination-weighted policy updating, with the corresponding mapping between scores and weights provided in Table~\ref{app_table1}.
Additionally, GPT-4V is required to categorize sentences with incorrect description as either \textit{image recognition errors} or \textit{language comprehension errors}. The classification results $S_\mathrm{img}$ is utilized for image-weighted policy updating, as defined in Eq.~\eqref{eq8}. Table~\ref{app_table2} outlines the mapping between these classifications and their respective updating weights.
Note that both $S_\mathrm{hal}$ and $S_\mathrm{img}$ are evaluated at the sentence level, ensuring that $W_\mathrm{hal}$ and $W_\mathrm{img}$ are assigned the same value for each token within a sentence.
Lastly but most importantly, GPT-4V is also instructed to make minimal revisions to any erroneous sentences, and the aggregate of these revised sentences is denoted as $\mathbf{y}_\mathrm{Rev}$. Please refer to Chapter~\ref{app_prompt} for detailed prompt and example. In our implementation, we utilize the GPT-4V version from 2024-02-15, with the generation temperature set to 0.

\begin{table}[ht]
\caption{GPT-4V assigned hallucination scores and the corresponding update weights for language correction loss, as described in Eq.~\eqref{eq7}.}\label{app_table1}
\vspace{-2mm}
\small
\centering
\begin{tabular}{ccc}
\hline
\begin{tabular}[c]{@{}c@{}}Hallucination\\ Severity \end{tabular} & \begin{tabular}[c]{@{}c@{}}Score from GPT-4V \\ ($S_\mathrm{hal}$) \end{tabular} & \begin{tabular}[c]{@{}c@{}}Updating Weight \\( $W_\mathrm{hal}$) \end{tabular} \\ \hline
Not at all & 4 & 2.5 \\
Minor & 3 & 2.0 \\
Major & 2 & 1.5 \\
Totally & 1 & 1.0 \\ \hline
\end{tabular}
\end{table}

\begin{table}[ht]
\caption{GPT-4V labeled error types and the corresponding update weights for image focus loss, as described in Eq.~\eqref{eq8}.}\label{app_table2}
\vspace{-2mm}
\small
\centering
\begin{tabular}{lc}
\hline
\begin{tabular}[c]{@{}c@{}}Label from GPT-4V\\ ($S_\mathrm{img}$)\end{tabular} & \begin{tabular}[c]{@{}c@{}}Updating Weight\\ ($W_\mathrm{img}$)\end{tabular} \\ \hline
correct & 1.0 \\
language\_comprehension\_error & 1.0 \\
image\_recognition\_error & 3.0 \\ \hline
\end{tabular}
\end{table}

\definecolor{hl_blue}{HTML}{70F3FF}
\definecolor{hl_yellow}{HTML}{FFFF00}
\definecolor{hl_green}{HTML}{29FC2A}
\begin{figure*}[t]
  \centering
   \includegraphics[width=0.99\linewidth]{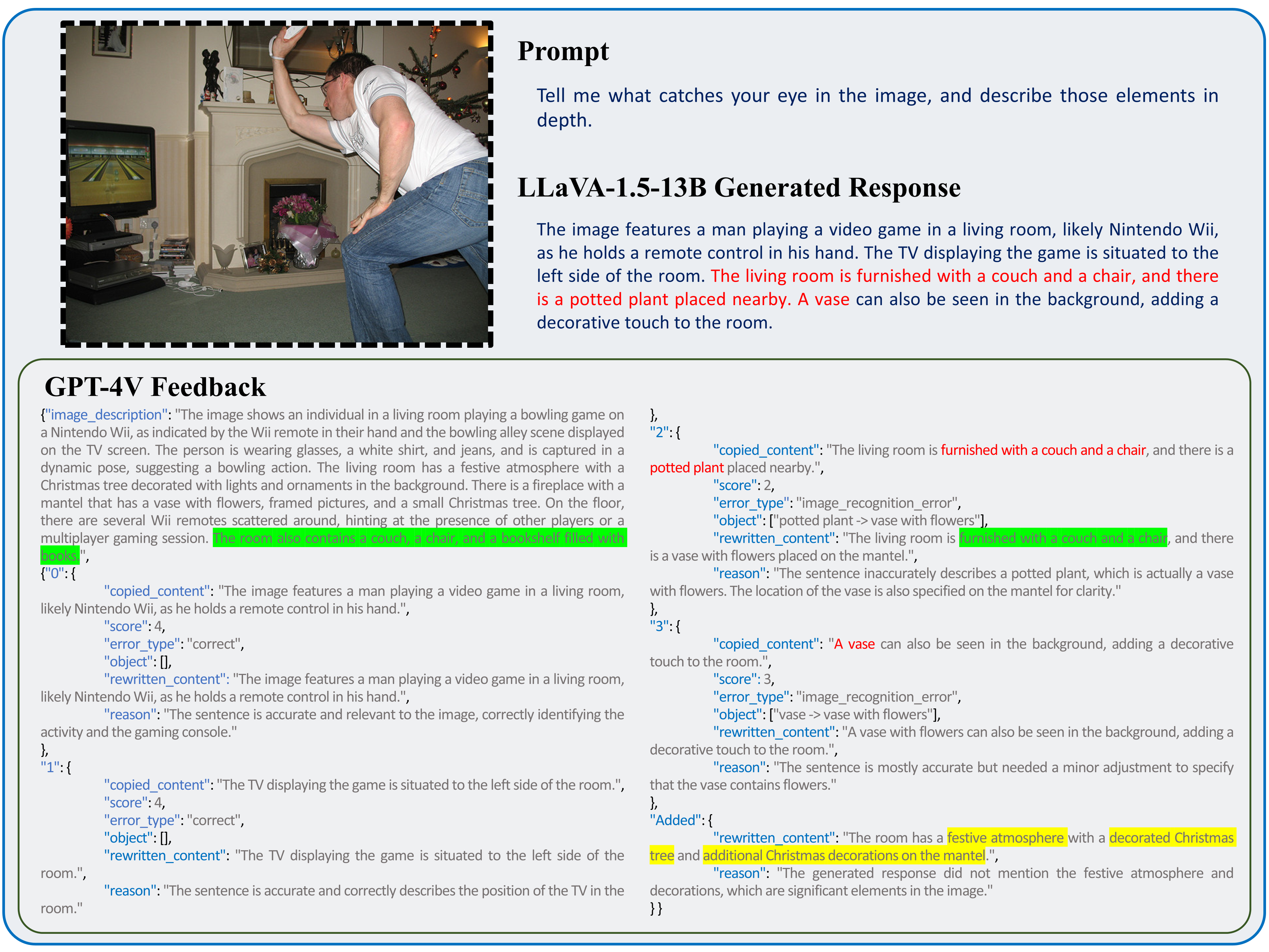}
   \vspace{-2mm}
   \caption{Example of feedback from GPT-4V. Hallucinated parts in the base-model generated responses are marked in \textcolor{red}{red}, missing details are {\sethlcolor{hl_yellow}\hl{highlighted in yellow}}. Note that the feedback from GPT-4V also contain hallucinations, as {\sethlcolor{hl_green}\hl{highlighted in green}}.}\label{fig_app1}
\end{figure*}

After completing the data collection, we proceed with a two-phase training for the initial models as detailed in Algorithm~\ref{alg_opadpo}. The first phase training (line 1-7) termed On-Policy Alignment (OPA), involves performing a 2-epoch LoRA-SFT on both ground-truth responses and GPT-4V revised responses.
The entire backbone model, including the vision encoder and multimodal connection layers, is wrapped with LoRA modules. We employ a cosine learning rate schedule beginning at 2e-5 with a batch size of 128. The LoRA rank is set to 256, and LoRA alpha is set to 512. The updated policy from this phase is denoted as $\pi_\mathrm{OPA}$, which serves as the initial (reference) policy for the subsequent OPA-DPO training.
The second phase of training (lines 8-18) uses the same LoRA module as in phase 1, extending over 4 additional epochs with a batch size of 32 and a cosine learning rate starting at 1e-6.
In our equations, we set $\beta=0.1$ in Eqs.~\eqref{eq7}\eqref{eq8}\eqref{eq9}, $\delta=0$ in Eq.~\eqref{eq9}, and $\gamma_1=0.2, \gamma_2=1.0$ in Eq.~\eqref{eq10}.
For the distorted images $\mathbf{m'}$ in Eq.~\eqref{eq8}, we randomly mask 30\% of pixels, assigning the masked areas the average pixel values. For ablation studies on the relative hyperparameters, please refer to Chapter~\ref{sec_add_ablation}.

\subsection{Prompts for GPT-4V}\label{app_prompt}

To obtain fine-grained feedback from GPT-4V, we crafted a detailed prompt, as shown in the TextBox on the following page. To establish a one-to-one correspondence between the revised and original responses, we instruct GPT-4V to first copy the generated sentence before proceeding with assessment and revision. Additionally, we request that GPT-4V provide the rationale behind its assigned score or revision. It is important to note that GPT-4V may itself produce hallucinations, which can affect the reliability of its feedback. An example is provided in Figure~\ref{fig_app1}.

{
\onecolumn
\footnotesize
\noindent\fbox{%
    \parbox{\textwidth}{%
    \begin{center}
        \normalsize \textbf{GPT-4V Prompt for Fine-Grained Sentence-Level Revision of Generated Responses}
    \end{center}
        Your role is as a discerning assistant tasked with evaluating and refining responses for multimodal tasks. Upon being presented with a question that requires the interpretation of both text and images, you will receive two distinct responses. The first is crafted by our sophisticated multimodal model, while the second represents an approximate ideal answer—it may be incomplete or incorrect. You will also be provided with the images pertinent to the question.

        Your objective is to meticulously assess these responses. You are to enhance the model-generated response by making precise, minimal modifications that bring it into closer alignment with both the image and the approximate ideal answer. Your revisions should preserve the integrity of the original response as much as possible.

        Be mindful that the approximate ideal response may not contain all the necessary information to fully address the question or may include mistakes. In such cases, you must carefully evaluate the accuracy of the model-generated response by consulting the image, which serves as the primary reference.

        Your analysis should prioritize the information provided in the image to ascertain the accuracy and completeness of the model-generated response. The ultimate goal is to ensure that the final response is both accurate in relation to the images and as informative as possible while remaining true to the content originally produced by the model.\\

        Your task involves meticulous scrutiny of the generated response to a multimodal task, sentence by sentence. Here's how you should approach the revision process:
        
        \hspace{1em} Evaluate each sentence within the generated response.
        
        \hspace{2em} - If a sentence is both accurate and relevant to the task, it should remain unchanged.

        \hspace{2em} - If you encounter a sentence that is only partially correct, carefully adjust the erroneous or incomplete segments to improve its precision. Ensure that these modifications are minimal and directly address the inaccuracies.

        \hspace{2em} - If you find any sentences that contain hallucinations or extraneous information, these must be either rephrased or replaced entirely. Use the image and the approximate ideal response as your sources for correction, aiming to retain the essence of the original content when possible.\\

        You are to present your output in a structured JSON format. Begin with the key ``image\_description" where a comprehensive description of the provided images should be articulated. Following this, evaluate the generated response sentence by sentence. For each sentence, craft a JSON object that contains the original sentence, your refined version, and a brief commentary explaining your revisions. The format is as follows:

        \hspace{1em} 1. ``copied\_content": Copy and paste the original sentence as it appears in the generated response.

        \hspace{1em} 2. ``score": Provide a score between 1 and 4, reflecting the sentence's accuracy and relevance to the image and question:
        
        \hspace{2em} - 4 for a sentence that is completely accurate and relevant, aligning perfectly with the image information and the approximate ideal answer, requiring no adjustments.
        
        \hspace{2em} - 3 for a sentence that is largely correct but needs minor tweaks, like an accurate object described with an incorrect count or size.
        
        \hspace{2em} - 2 for a sentence with substantial issues requiring significant changes, such as incorrect object recognition or incorrect relationships between objects.
        
        \hspace{2em} - 1 for a sentence that is completely irrelevant or incorrect, with no relation to the image or the question at hand.

        \hspace{1em} 3. ``error\_type": Specify the type of error detected in the sentence:

        \hspace{2em} - ``correct" if the sentence is accurate or requires only minor adjustments, applicable only to a score of 4.
        
        \hspace{2em} - ``image\_recognition\_error" when the error arises from an incorrect interpretation of the visual content, like mistaking an apple for a pear.
        
        \hspace{2em} - ``language\_comprehension\_error" when the image is correctly understood, but the language used is incorrect, such as placing the Eiffel Tower in Berlin instead of Paris.

        \hspace{1em} 4. ``object": List any objects that are hallucinated or misidentified, and provide the correct identification. Leave this field empty if there are no hallucinations or misidentifications.

        \hspace{2em} - For instance, if the sentence inaccurately identifies a cat sleeping on a table as a dog standing on a blanket, the ``object" should be [``dog -$>$  cat", ``standing -$>$ sleeping", ``blanket -$>$ table"].

        \hspace{1em} 5. ``rewritten\_content": Present the corrected sentence after applying necessary adjustments, considering all information from the image captions and the approximate ideal answer.

        \hspace{1em} 6. ``reason": Explain the rationale for the given score, the identified error type, and any modifications made. This should include the reasoning behind changes and the decision to maintain certain parts of the original sentence.\\
        
        If the rewritten sentences still lack essential information necessary for answering the given questions, add the missing part to the ``Added" section and incorporate that missing information minimally. Only do this if absolutely necessary.
        
        You should never bring other hallucinations into the rewritten parts. Only do the modifications when you are one hundred percent sure that the original sentence is incorrect or irrelevant.
        Please note that the rewritten sentence should retain as much of the generated response as possible. All unnecessary changes should be minimized.

    }%
}
}

\twocolumn
\section{Additional Experiments}\label{sec_add_exps}

\subsection{Helpfulness Benchmark Evaluations.}\label{sec_add_llavabench}

To demonstrate that the exceptional performance of OPA-DPO on hallucination-related metrics does not result in a decline in helpfulness-related metrics, we evaluated the performance of various RLHF/RLAIF-based algorithms designed to enhance LVLMs on the LLaVA-Bench \cite{liu2024llava}, as shown in Table~\ref{table_llava_bench}. The results indicate that the OPA-DPO trained model performs at an upper-middle level. With the exception of LLaVA-RLHF, the performance of each algorithm on the LLaVA-Benchmark shows minimal variation. However, LLaVA-RLHF is significantly less effective than other algorithms in hallucination-related metrics.

\begin{table}[ht]
\footnotesize
\caption{Comparison of RLAIF/RLHF-based algorithms for enhancing LVLMs on LLaVA-Bench.}\label{table_llava_bench}
\vspace{-2mm}
\scriptsize
\resizebox{0.47\textwidth}{22mm}{
\begin{tabular}{lcccc}
\hline
\rowcolor[HTML]{EFEFEF} 
\multicolumn{1}{c|}{\cellcolor[HTML]{EFEFEF}} & \multicolumn{4}{c}{\cellcolor[HTML]{EFEFEF}\textbf{LLaVA-Bench}} \\ \cline{2-5} 
\rowcolor[HTML]{EFEFEF} 
\multicolumn{1}{c|}{\multirow{-2}{*}{\cellcolor[HTML]{EFEFEF}\textbf{Algorithm}}} & \textbf{Conv.↑} & \textbf{Detail↑} & \textbf{Comp.↑} & \textbf{All↑} \\ \hline \addlinespace[0.5mm]
\textbf{LLaVA-Instruct-1.5-7B \cite{liu2024llava, liu2024llava2}} & 84.1 & 74.4 & 89.8 & 83.0 \\
+ LLaVA-RLHF \cite{sun2023llava_rlhf}& 84.1 & 75.3 & 106.8 & 88.9 \\
+ HA-DPO \cite{zhao2023hadpo}& 80.7 & 74.5 & 88.4 & 81.4 \\
+ POVID \cite{zhou2024povid}& 84.9 & 77.3 & 90.3 & 84.3 \\
+ RLAIF-V \cite{yu2024rlaif}& 75.8 & 83.7 & 90.7 & 83.5 \\
\rowcolor[HTML]{ECF4FF} 
\textbf{+ OPA-DPO (ours)} & 82.1 & 79.5 & 87.9 & 83.2 \\ \addlinespace[0.5mm] \hline \addlinespace[0.5mm]
\textbf{LLaVA-Instruct-1.5-13B \cite{liu2024llava, liu2024llava2}} & 79.6 & 77.3 & 91.4 & 82.9 \\
+ LLaVA-RLHF \cite{sun2023llava_rlhf}& 93.1 & 76.2 & 105.6 & 91.8 \\
+ RLHF-V \cite{yu2024rlhfv}& 93.1 & 75.3 & 91.6 & 86.7 \\
+ HSA-DPO \cite{xiao2024hsadpo}& 76.0 & 71.8 & 88.2 & 80.5 \\
\rowcolor[HTML]{ECF4FF} 
\textbf{+ OPA-DPO (ours)} & 87.1 & 78.3 & 90.7 & 85.5 \\ \addlinespace[0.5mm] \hline
\end{tabular}
}
\end{table}

\subsection{Additional Ablation Studies.}\label{sec_add_ablation}

\begin{table}[ht]
\footnotesize
\caption{Ablation studies on the mask ratio of the distorted image and the term coefficient $\gamma_1$ in the image focus mechanism.}\label{table_IF_ablation}
\vspace{-2mm}
\scriptsize
\resizebox{0.47\textwidth}{22mm}{
\setlength{\tabcolsep}{1pt}
\begin{tabular}{ccc|ccc|cc|cc}
\hline
\rowcolor[HTML]{EFEFEF} 
\cellcolor[HTML]{EFEFEF} & \cellcolor[HTML]{EFEFEF} & \cellcolor[HTML]{EFEFEF} & \multicolumn{3}{c|}{\cellcolor[HTML]{EFEFEF}\textbf{AMBER}} & \multicolumn{2}{c|}{\cellcolor[HTML]{EFEFEF}\textbf{MMHal-Bench}} & \multicolumn{2}{c}{\cellcolor[HTML]{EFEFEF}\textbf{Object Hal}} \\
\rowcolor[HTML]{EFEFEF} 
\multirow{-2}{*}{\cellcolor[HTML]{EFEFEF}\textbf{\begin{tabular}[c]{@{}c@{}}Model\\ Size\end{tabular}}} & \multirow{-2}{*}{\cellcolor[HTML]{EFEFEF}\textbf{\begin{tabular}[c]{@{}c@{}}Mask\\ Ratio\end{tabular}}} & \multirow{-2}{*}{\cellcolor[HTML]{EFEFEF}\textbf{\begin{tabular}[c]{@{}c@{}}IF Coef\\ $\gamma_1$\end{tabular}}} & \textbf{Cover↑} & \textbf{HalRate↓} & \textbf{repeat} & \textbf{Score↑} & \textbf{HalRate↓} & \textbf{CHAIRs↓} & \textbf{CHAIRi↓} \\ \hline \addlinespace[0.5mm]
 & 0.1 & 0.2 & 46.1 & 12.5 & 0.3\% & 2.60 & 0.49 & 14.67 & 4.28 \\
 & \cellcolor[HTML]{ECF4FF}0.3 & \cellcolor[HTML]{ECF4FF}0.2 & \cellcolor[HTML]{ECF4FF}47.9 & \cellcolor[HTML]{ECF4FF}11.6 & \cellcolor[HTML]{ECF4FF}0.6\% & \cellcolor[HTML]{ECF4FF}2.83 & \cellcolor[HTML]{ECF4FF}0.45 & \cellcolor[HTML]{ECF4FF}13.00 & \cellcolor[HTML]{ECF4FF}4.25 \\
 & 0.5 & 0.2 & 46.3 & 11.6 & 0.3\% & 2.73 & 0.47 & 14.67 & 4.18 \\
 & 0.7 & 0.2 & 45.6 & 12.2 & 0.2\% & 2.69 & 0.47 & 13.33 & 4.45 \\
 & 0.3 & 0.1 & 46.2 & 12.3 & 0.6\% & 2.70 & 0.47 & 14.67 & 4.05 \\
 & 0.3 & 0.5 & 44.3 & 11.1 & 5.3\% & 2.79 & 0.45 & 14.67 & 4.32 \\
\multirow{-7}{*}{\textbf{7B}} & 0.3 & 1.0 & 43.5 & 9.3 & 20.8\% & 2.26 & 0.59 & 9.67 & 2.98 \\ \addlinespace[0.5mm] \hline \addlinespace[0.5mm]
 & 0.1 & 0.2 & 48.3 & 13.9 & 0.4\% & 2.84 & 0.45 & 19.00 & 6.16 \\
 & \cellcolor[HTML]{ECF4FF}0.3 & \cellcolor[HTML]{ECF4FF}0.2 & \cellcolor[HTML]{ECF4FF}48.3 & \cellcolor[HTML]{ECF4FF}12.8 & \cellcolor[HTML]{ECF4FF}0.8\% & \cellcolor[HTML]{ECF4FF}3.07 & \cellcolor[HTML]{ECF4FF}0.39 & \cellcolor[HTML]{ECF4FF}16.33 & \cellcolor[HTML]{ECF4FF}5.48 \\
 & 0.5 & 0.2 & 48.2 & 13.4 & 0.8\% & 2.99 & 0.41 & 18.00 & 5.41 \\
 & 0.7 & 0.2 & 48.3 & 13.9 & 0.4\% & 2.84 & 0.45 & 19.00 & 6.16 \\
 & 0.3 & 0.1 & 48.3 & 13.0 & 0.2\% & 2.95 & 0.42 & 18.33 & 5.89 \\
 & 0.3 & 0.5 & 48.1 & 12.3 & 2.6\% & 2.97 & 0.44 & 16.67 & 5.35 \\
\multirow{-7}{*}{\textbf{13B}} & 0.3 & 1.0 & 46.1 & 10.3 & 7.9\% & 2.72 & 0.44 & 16.33 & 5.00 \\ \addlinespace[0.5mm] \hline
\end{tabular}
}
\end{table}

As an important component of OPA-DPO, the image focus mechanism (see Chapter~\ref{OPA_DPO}) involves two hyperparameters that require tuning: the term coefficient $\gamma_1$ in Eq.~\ref{eq10} and the mask ratio of the distorted image. We found that setting $\gamma_1$ to 0.2 and randomly masking 30\% of the pixels is optimal for both the LLaVA-1.5-7B and LLaVA-1.5-13B models. In contrast, the pioneering algorithm mDPO \cite{wang2024mdpo}, which first utilized this mechanism, opts to set $\gamma_1$ to 1.0 and employs a variable masking ratio of 0-20\% of pixels randomly. 
We find that the mask ratio has a slight impact on the model's performance, whereas the term coefficient $\gamma_1$ has a more significant effect. 
In particular, setting the coefficient too high results in excellent performance in metrics related to hallucination rate, but at the cost of being overly conservative and severely lacking in explanatory detail. Additionally, the model tends to repeat its last sentence or words and fails to generate an EOS token when using greedy sampling. As a compromise, we set $\gamma_1=0.2$.
Ablation studies supporting our findings are presented in Table~\ref{table_IF_ablation}.

\section{Additional Qualitative Examples}\label{sec_qual_results}

\paragraph{Image Descriptions.}
As introduced in Chapter~\ref{sec5_exp}, OPA-DPO is particularly effective in preventing hallucinations by adopting a somewhat conservative strategy that avoids uncertain assertions. Such strategy significantly enhances the credibility of the responses but may lead to the omission of some ambiguous details, necessitating a trade-off. In addition to the case presented in Chapter~\ref{sec5_5case_study}, we offer further examples involving image detail descriptions, as illustrated in Figures~\ref{app_case1}, \ref{app_case2}, \ref{app_case3}, and \ref{app_case4}. In these cases, the initial model's output contained numerous hallucinations and flawed reasoning. This issue persisted even after training with naive DPO without OPA. However, after applying OPA to 4.8k samples, hallucinations were nearly eliminated, with only minor instances remaining. The subsequent implementation of OPA-DPO completely resolved these issues, although some details from the original description were omitted. It is important to note that the omitted details are often not central to the image's main information and do not cause the overall description to deviate.

\paragraph{False Premise Queries.}
Another interesting phenomenon we observed in our experiments is that, LVLMs consistently experience hallucinations when presented with queries based on false premises.
These queries contain objects or details that do not exist in the image or are irrelevant to it.
For example, the LVLM is asked to describe the girl's outfit given a picture of a basketball. 
As demonstrated in Figures \ref{app_case5}, \ref{app_case6}, and \ref{app_case7}, the base model consistently produces absurd responses to nonsensical questions due to linguistic inertia. 
The application of DPO without OPA does not generally modify these responses. 
Furthermore, utilizing the OPA operation in isolation is sometimes insufficient to address the issue. 
However, when both methods are combined, through training with OPA-DPO, the model is able to discern false premises in queries or prompts and provide reasoned responses.

\definecolor{hl_blue}{HTML}{70F3FF}
\definecolor{hl_yellow}{HTML}{FFFF00}

\begin{figure*}[ht]
  \centering
   \includegraphics[width=0.99\linewidth]{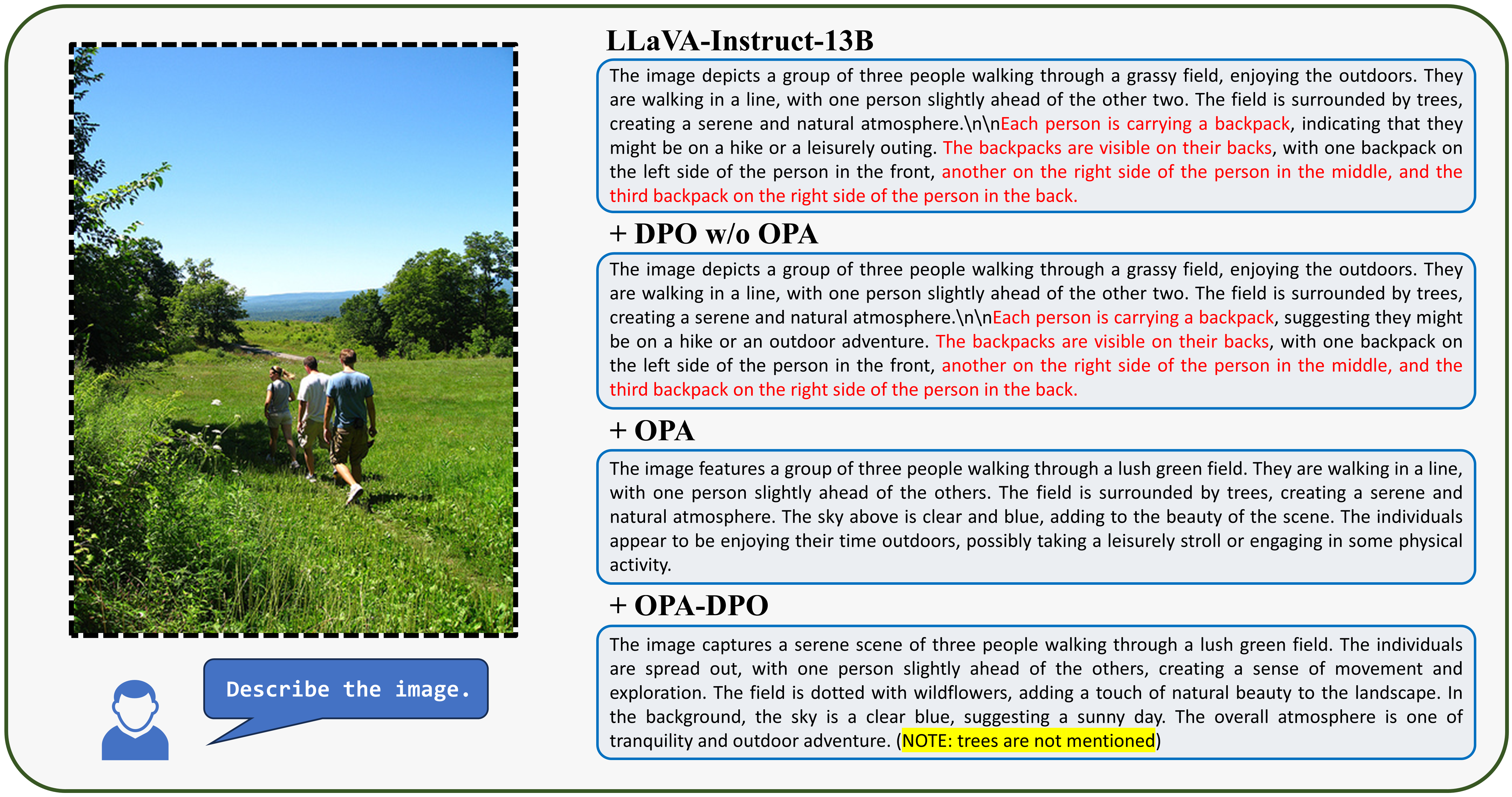}
   \vspace{-2mm}
   \caption{Qualitative results of different models. Hallucinated parts are marked in \textcolor{red}{red}, and missing details are {\sethlcolor{hl_yellow}\hl{highlighted in yellow}}.}\label{app_case1}
   \vspace{-5mm}
\end{figure*}

\begin{figure*}[ht]
  \centering
   \includegraphics[width=0.99\linewidth]{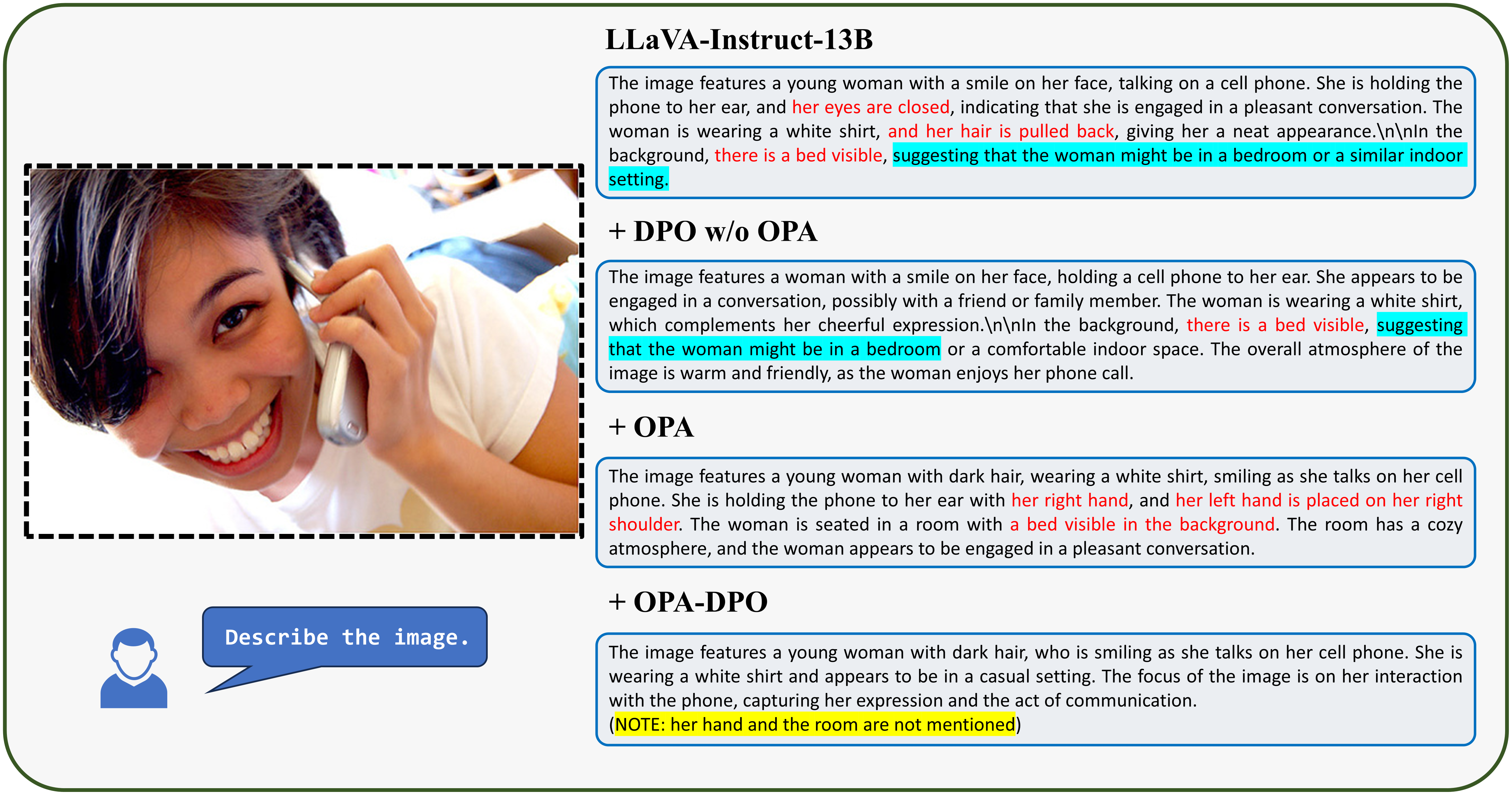}
   \vspace{-2mm}
   \caption{Qualitative results of different models. Hallucinated parts are marked in \textcolor{red}{red}, flawed reasoning is {\sethlcolor{hl_blue}\hl{highlighted in blue}}, and missing details are {\sethlcolor{hl_yellow}\hl{highlighted in yellow}}.}\label{app_case2}
   \vspace{-5mm}
\end{figure*}

\begin{figure*}[ht]
  \centering
   \includegraphics[width=0.99\linewidth]{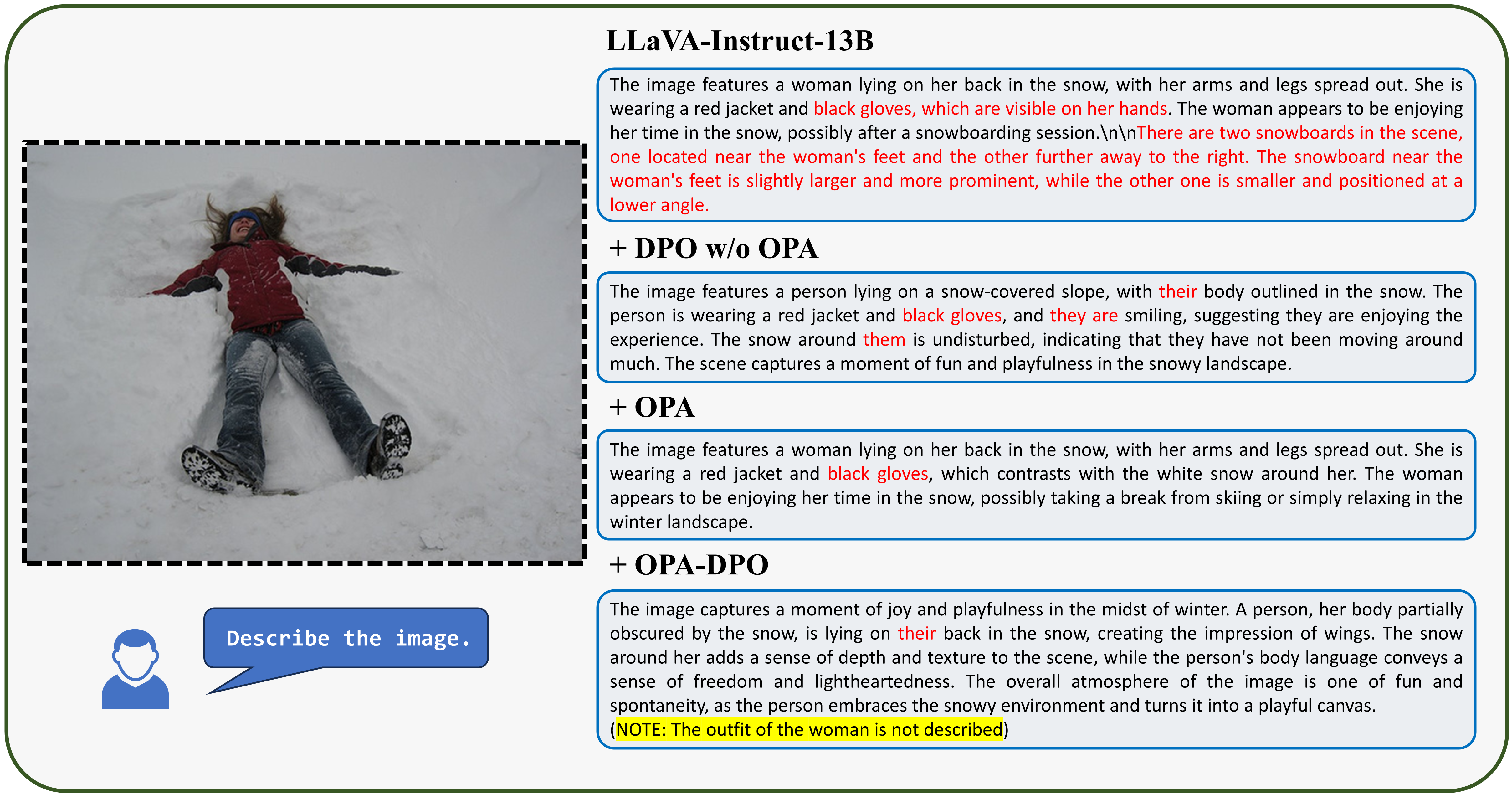}
   \vspace{-2mm}
   \caption{Qualitative results of different models. Hallucinated parts are marked in \textcolor{red}{red}, and missing details are {\sethlcolor{hl_yellow}\hl{highlighted in yellow}}.}\label{app_case3}
   \vspace{-5mm}
\end{figure*}

\begin{figure*}[ht]
  \centering
   \includegraphics[width=0.99\linewidth]{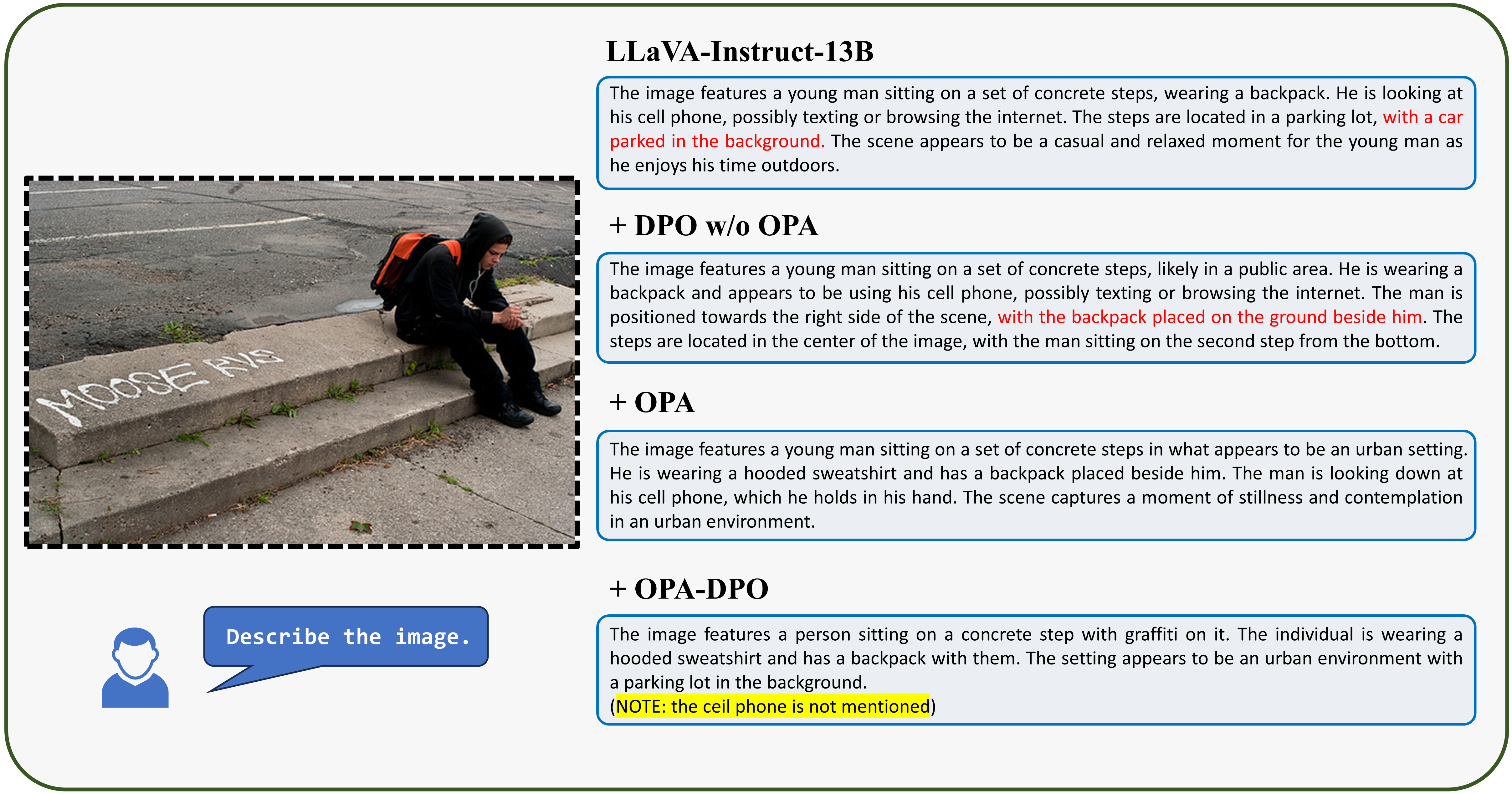}
   \vspace{-2mm}
   \caption{Qualitative results of different models. Hallucinated parts are marked in \textcolor{red}{red}, and missing details are {\sethlcolor{hl_yellow}\hl{highlighted in yellow}}.}\label{app_case4}
   \vspace{-5mm}
\end{figure*}

\begin{figure*}[ht]
  \centering
   \includegraphics[width=0.99\linewidth]{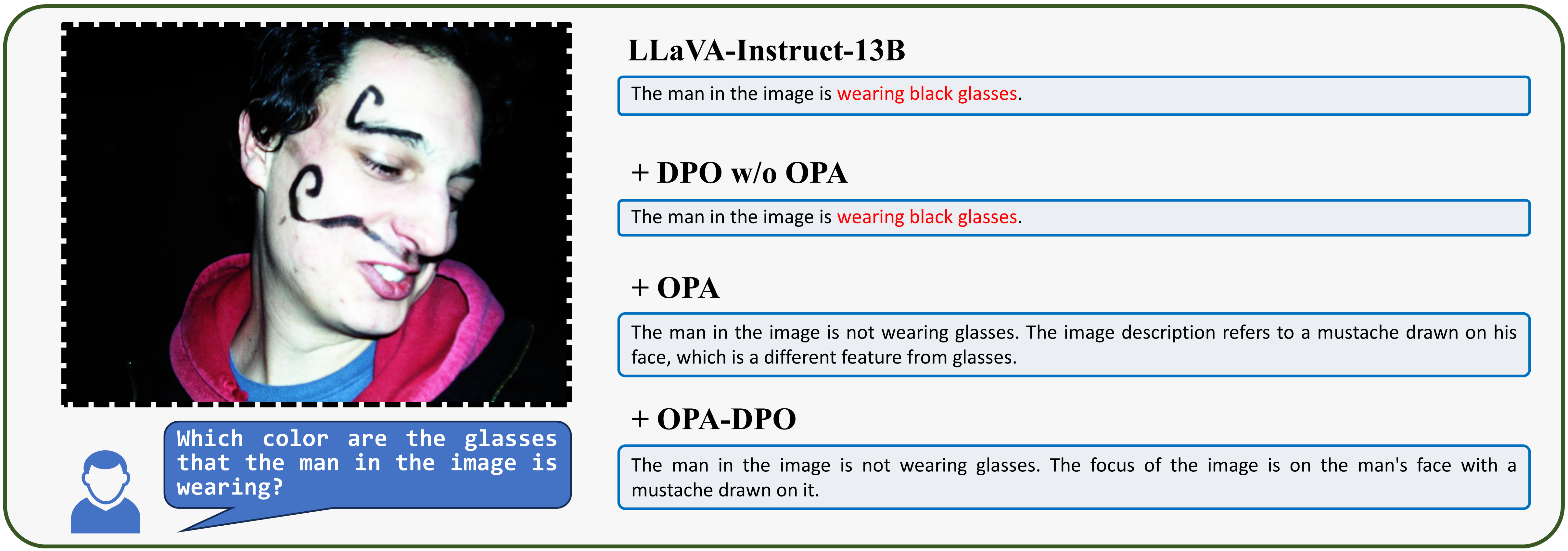}
   \vspace{-2mm}
   \caption{Qualitative results of different models. Hallucinated parts are marked in \textcolor{red}{red}.}\label{app_case5}
   \vspace{-5mm}
\end{figure*}

\begin{figure*}[ht]
  \centering
   \includegraphics[width=0.99\linewidth]{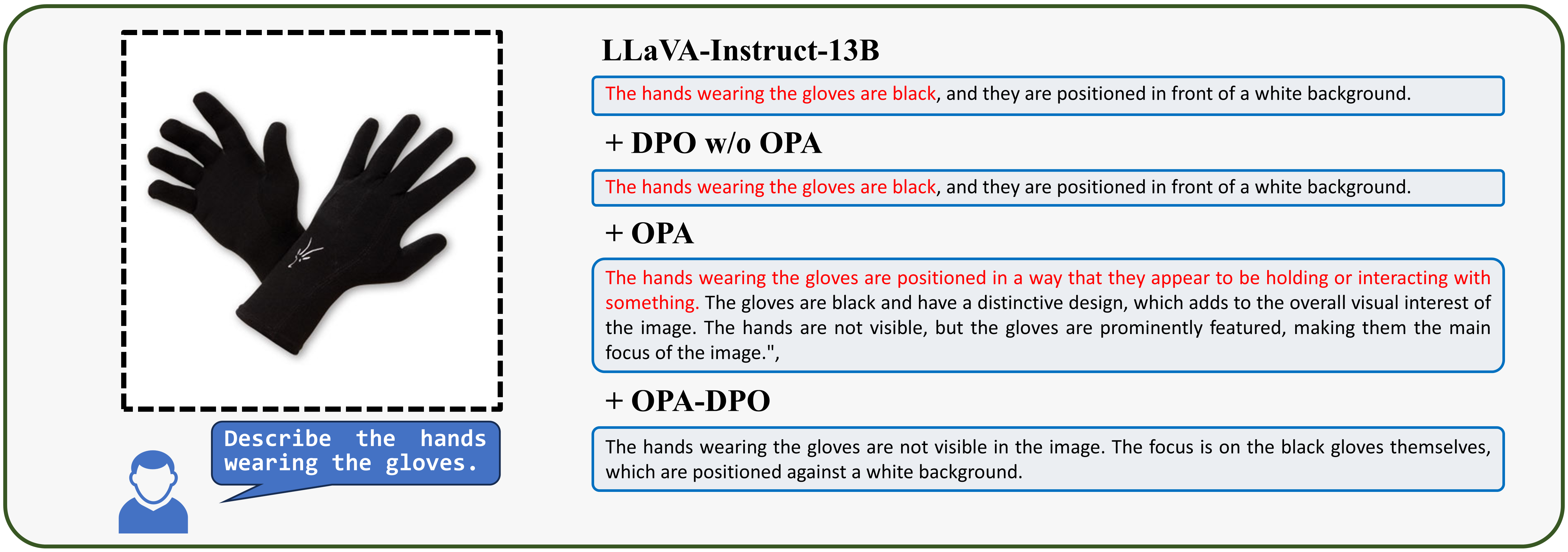}
   \vspace{-2mm}
   \caption{Qualitative results of different models. Hallucinated parts are marked in \textcolor{red}{red}.}\label{app_case6}
   \vspace{-5mm}
\end{figure*}

\begin{figure*}[ht]
  \centering
   \includegraphics[width=0.99\linewidth]{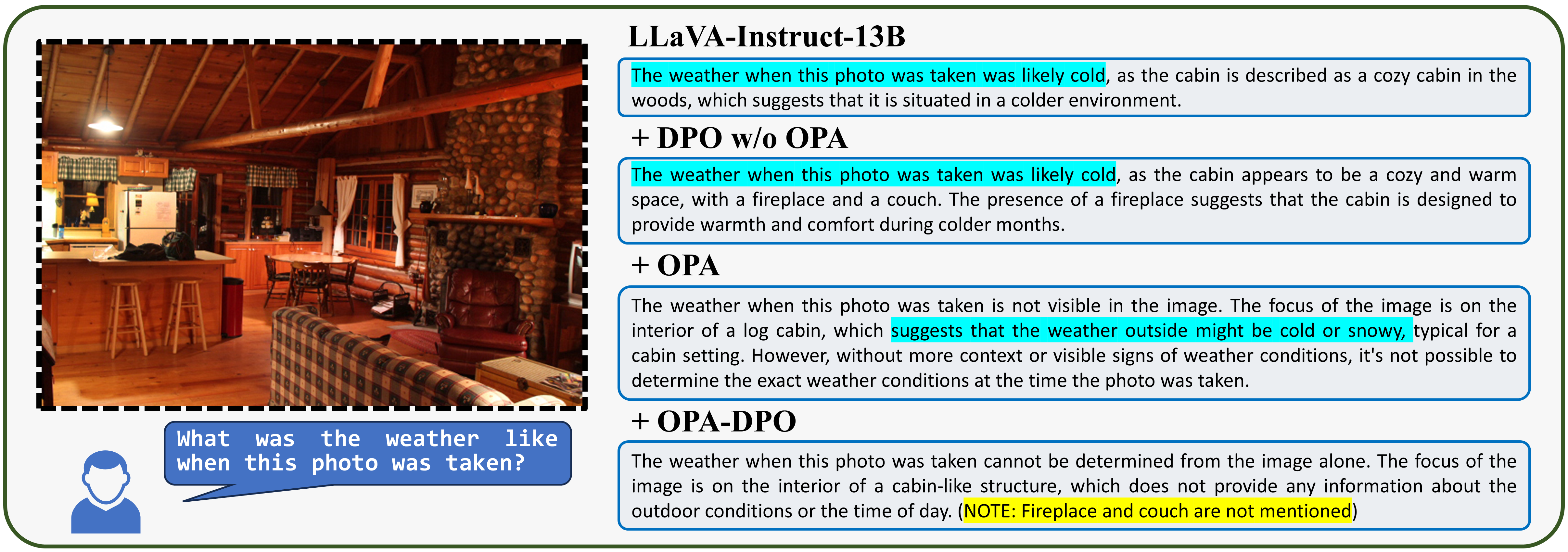}
   \vspace{-2mm}
   \caption{Qualitative results of different models. Flawed reasoning is {\sethlcolor{hl_blue}\hl{highlighted in blue}}, and missing details are {\sethlcolor{hl_yellow}\hl{highlighted in yellow}}.}\label{app_case7}
   \vspace{-5mm}
\end{figure*}
}

\end{document}